%% file: neurips_main.tex
\def\year{2023}\relax
\documentclass[letterpaper]{article} 
\usepackage{aaai23}  
\usepackage{times}  
\usepackage{helvet}  
\usepackage{courier}  
\usepackage[hyphens]{url}  
\usepackage{graphicx} 
\usepackage{natbib}  
\usepackage{caption} 
\DeclareCaptionStyle{ruled}{labelfont=normalfont,labelsep=colon,strut=off} 
\usepackage{floatrow}
\usepackage{float}
\newfloatcommand{capbtabbox}{table}[][\FBwidth]

\usepackage[utf8]{inputenc} 
\usepackage[T1]{fontenc}    
\usepackage{amsfonts}       
\usepackage{nicefrac}       
\usepackage{microtype}      
\usepackage{algpseudocode}
\usepackage[ruled,vlined]{algorithm2e}
\usepackage{amsmath,bm}
\usepackage{amsthm}
\usepackage{bbold}
\usepackage{subfigure}
\usepackage{color}
\usepackage{xcolor}  
\usepackage{multirow}
\usepackage{balance}
\usepackage{enumitem}
\usepackage{wrapfig}
\usepackage{booktabs} 
\usepackage[english]{babel}

\DeclareMathOperator*{\argmin}{arg\,min}

\usepackage{xcolor}         
\algnewcommand{\LeftComment}[1]{ \(\triangleright\) #1}

\title{Eliciting Structural and Semantic Global Knowledge in \\ Unsupervised Graph Contrastive Learning}

\author {
    Kaize Ding\thanks{Indicates equal contribution.},
    Yancheng Wang\footnotemark[1],
    Yingzhen Yang,
    \textrm{and} Huan Liu
}
\affiliations {
    Arizona State University\\
    School of Computing and Augmented Intelligence\\
    kaize.ding@asu.edu, yancheng.wang@asu.edu, yingzhen.yang@asu.edu, huan.liu@asu.edu
}

\begin{document}

\maketitle

\begin{abstract}



Graph Contrastive Learning (GCL) has recently drawn much research interest for learning generalizable node representations in a self-supervised manner. In general, the contrastive learning process in GCL is performed on top of the representations learned by a graph neural network (GNN) backbone, which transforms and propagates the node contextual information based on its local neighborhoods. However, nodes sharing similar characteristics may not always be geographically close, which poses a great challenge for unsupervised GCL efforts due to their inherent limitations in capturing such global graph knowledge. In this work, we address their inherent limitations by proposing a simple yet effective framework -- \textit{\underline{S}imple Neural Networks with \underline{S}tructural and \underline{S}emantic \underline{C}ontrastive \underline{L}earning} (S$^3$-CL). Notably, by virtue of the proposed structural and semantic contrastive learning algorithms, even a simple neural network can learn expressive node representations that preserve valuable global structural and semantic patterns. Our experiments demonstrate that the node representations learned by S$^3$-CL achieve superior performance on different downstream tasks compared with the state-of-the-art unsupervised GCL methods. Implementation and more experimental details are publicly available
at \url{https://github.com/kaize0409/S-3-CL.}

\end{abstract}

\section{Introduction}

Learning expressive node representations of graph-structured data plays an essential role in a variety of real-world applications, ranging from social network analysis~\cite{kipf2017semi}, to drug discovery~\cite{protein_interface}, to financial fraud detection~\cite{ding2019deep}. Recently, graph neural networks (GNNs), which generally follow a recursive message-passing scheme, have emerged as powerful architectures in graph machine learning~\cite{kipf2017semi,gat,hamilton2017inductive,sgc,ding2020graph,wang2020next}. Though GNNs are empirically effective in handling supervised or semi-supervised graph machine learning tasks, the labor-intensive and resource-expensive data labeling cost is meanwhile unbearable~\cite{ding2022data,zhang2022few,ding2022toward}. To relieve the burdensome reliance on human-annotated labels, unsupervised (self-supervised) node representation learning with GNNs has drawn much research attention lately~\cite{kipf_2016_arxiv,velickovic_2019_iclr,you2020graph}.


More recently, contrastive learning~\cite{moco,simclr} has been actively explored to advance the performance of GNNs in graph self-supervised learning~\cite{velickovic_2019_iclr,you2020graph,hassani2020contrastive,qiu2020gcc,zhu2020graph}. In general, graph contrastive learning (GCL) methods learn representations by creating two augmented views of each graph element and maximizing the agreement between the encoded representations of the two augmented views. Correspondingly, the relevant view pairs (positive) will be pulled together, and the irrelevant view pairs (negative) will be pushed away in the latent space. With only non-semantic labels, unsupervised GCL can provide generalizable node representations for various downstream tasks~\cite{you2020graph,hassani2020contrastive,du2021graphgt}, becoming a prevailing paradigm in unsupervised node representation learning. 

Despite the success, the research of unsupervised GCL is still in its infancy -- most of the existing GCL methods learn node representations based on the information from the local neighborhoods due to the shallow property of conventional GNNs. While for real-world graphs, \textit{nodes sharing similar characteristics may not always be geographically close}, requiring the learning algorithm to retain such ``global'' awareness. However, it is a non-trivial task for the existing GCL methods built on top of shallow GNNs since they have inherent limitations in capturing either \textit{structural global knowledge} or \textit{semantic global knowledge}. Specifically: \textbf{(i)} from the structural perspective, long-range node interactions are highly desired for capturing structural global knowledge, especially for many downstream tasks that have large problem radii~\cite{alon2021bottleneck}. To this end, a straightforward way is to employ a deeper GNN encoder to encode the augmented graphs. However, directly stacking multiple GNN layers will not only lead to information distortion caused by the oversmoothing issue~\cite{chen2020measuring}, but also introduce additional training parameters that hamper the model training efficiency; and \textbf{(ii)} from the semantic perspective, existing unsupervised GCL methods predominately focus on instance-level contrast that leads to a latent space where all nodes are well-separated and each node is locally smooth~\cite{li2020prototypical} (i.e., input with different augmentations have similar representations), while the underlying semantic structure (i.e., intra-cluster compactness and inter-cluster separability) of the input graph is largely ignored~\cite{li2020prototypical}. The lack of prior knowledge of ground-truth labels (e.g., cluster/class numbers) leaves a significant gap for unsupervised GCL to consolidate the semantic structure from a global view in the latent space. Yet, how to bridge this gap remains unattended.

 In this paper, we address the aforementioned limitations by proposing a simple yet effective GCL framework, namely, S$^3$-CL (\textit{\underline{S}imple Neural Networks with \underline{S}tructural and \underline{S}emantic \underline{C}ontrastive \underline{L}earning}). The proposed two new contrastive learning algorithms enable the framework to outperform other GCL counterparts with a much simpler and parameter-less encoding backbone, such as an MLP or even a one-layer neural network. To capture long-range node interactions without oversmoothing, the \textit{structural contrastive learning} algorithm first generates multiple augmented views of the input graph based on different feature propagation scales (i.e., multi-scale feature propagation). Then by performing contrastive learning on the node representations learned from the local and multiple high-order views, the encoder network can improve node-wise discrimination by exploiting the consistency between the local and global structure information of each node. In the meantime, the \textit{semantic contrastive learning} algorithm further enhances intra-cluster compactness and inter-cluster separability to better consolidate the semantic structure from a global view. Specifically, it infers the clusters among nodes and their corresponding prototypes by a new Bayesian non-parametric algorithm and then performs semantic contrastive learning to enforce those nodes that are semantically similar to cluster around their corresponding cluster prototypes in the latent space. By jointly optimizing the structural and semantic contrastive losses, the pre-trained encoder network can learn highly expressive node representations for various downstream tasks without using any human-annotated labels. We summarize our contributions as follows:
\begin{itemize}[leftmargin=*,noitemsep,topsep=1.5pt]
    \item We develop a new GCL framework S$^3$-CL, which can learn expressive node representations in a self-supervised fashion by 
    using a simple and parameter-less encoding backbone. 
    
    \item We propose structural and semantic contrastive learning algorithms, which can be used for explicitly capturing the global structural and semantic patterns of the input graph.
    
    \item We conduct extensive experiments to show that our approach significantly outperforms the state-of-the-art GCL counterparts on various downstream tasks. 
\end{itemize}

\section{Preliminaries}
We start by introducing the notations used throughout the paper. An attributed graph with $N$ nodes can be formally represented by $\mathcal{G} = (\mathcal{V}, \mathcal{E}, \mathbf{X})$, where $\mathcal V = \{v_1, v_2, \dots, v_N\}$ and $\mathcal{E}\subseteq\mathcal{V}\times\mathcal{V}$ denote the set of nodes and edges respectively. Let $\mathbf{A} \in \{0, 1\}^{N \times N}$ be the adjacency matrix of graph $\mathcal{G}$. $\mathbf{A}_{ij} = 1$ if and only if $(v_i, v_j)\in \mathcal{E}$. $\Tilde{\mathbf{A}}$ stands for the adjacency matrix for a graph with added self-loops $\mathbf{I}$. We let $\mathbf{D}$ and $\Tilde{\mathbf{D}}$ denote the diagonal degree matrix of $\mathbf{A}$ and $\Tilde{\mathbf{A}}$ respectively. $\mathbf{x}_i$ is the $i$-th row of the attribute matrix $\mathbf{X} \in \mathbb{R}^{N \times D}$, which denotes the feature of node $v_i$. Hence, an attributed graph can also be described as $\mathcal{G} = (\mathbf{X}, \mathbf{A})$ for simplicity.

\noindent\textbf{Graph Contrastive Learning.} In general, graph contrastive learning aims to pre-train a graph encoder that can maximize the node-wise agreement between two augmented views of the same graph element in the latent space via a contrastive loss. Generally, given an attributed graph $\mathcal{G} = (\mathbf{X}, \mathbf{A})$, two different augmented views of the graph, denoted as $\mathcal{G}^{(1)} = (\mathbf{X}^{(1)}, \mathbf{A}^{(1)})$ and $\mathcal{G}^{(2)} = (\mathbf{X}^{(2)}, \mathbf{A}^{(2)})$, are generated through the data augmentation function(s). The node representations on $\mathcal{G}^{1}$ and $\mathcal{G}^{(2)}$ are denoted as $\mathbf{H}^{(1)}=f_{\bm\theta}(\mathbf{X}^{(1)}, \mathbf{A}^{(1)})$ and $\mathbf{H}^{(2)}=f_{\bm\theta}(\mathbf{X}^{(2)}, \mathbf{A}^{(2)})$, where $f_{\bm\theta}(\cdot)$ is an encoder network. The agreement between the node representations is commonly measured through Mutual Information (MI). Thus, the contrastive objective can be generally formulated as:
\begin{equation}
    \max_{\theta}\sum_{i=1}^{N}\mathcal{MI}(\mathbf{h}^{(1)}_i,\mathbf{h}^{(2)}_i).
\end{equation}

Following this formulation, Deep Graph Infomax (DGI)~\cite{velickovic_2019_iclr} is the first method that contrasts the patch representations with high-level graph representations by maximizing their mutual information.
MVGRL~\cite{hassani2020contrastive} adopts graph diffusion to generate an augmented view, and contrast representations of first-order neighbors with a graph diffusion. GCC~\cite{qiu2020gcc} and GRACE~\cite{Zhu:2020vf} create the augmented views by sampling subgraphs.
MERIT~\cite{jin2021multi} adopts a siamese self-distillation network and performs contrastive learning across views and networks at the same time. Nonetheless, existing unsupervised GCL methods only focus on short-range node interactions and are also ineffective in capturing the semantic structure of graphs. 


\section{Methodology}

\begin{figure*}[t]
  \centering
  \scalebox{0.9}{
    {
    \includegraphics[width=1.0\textwidth]{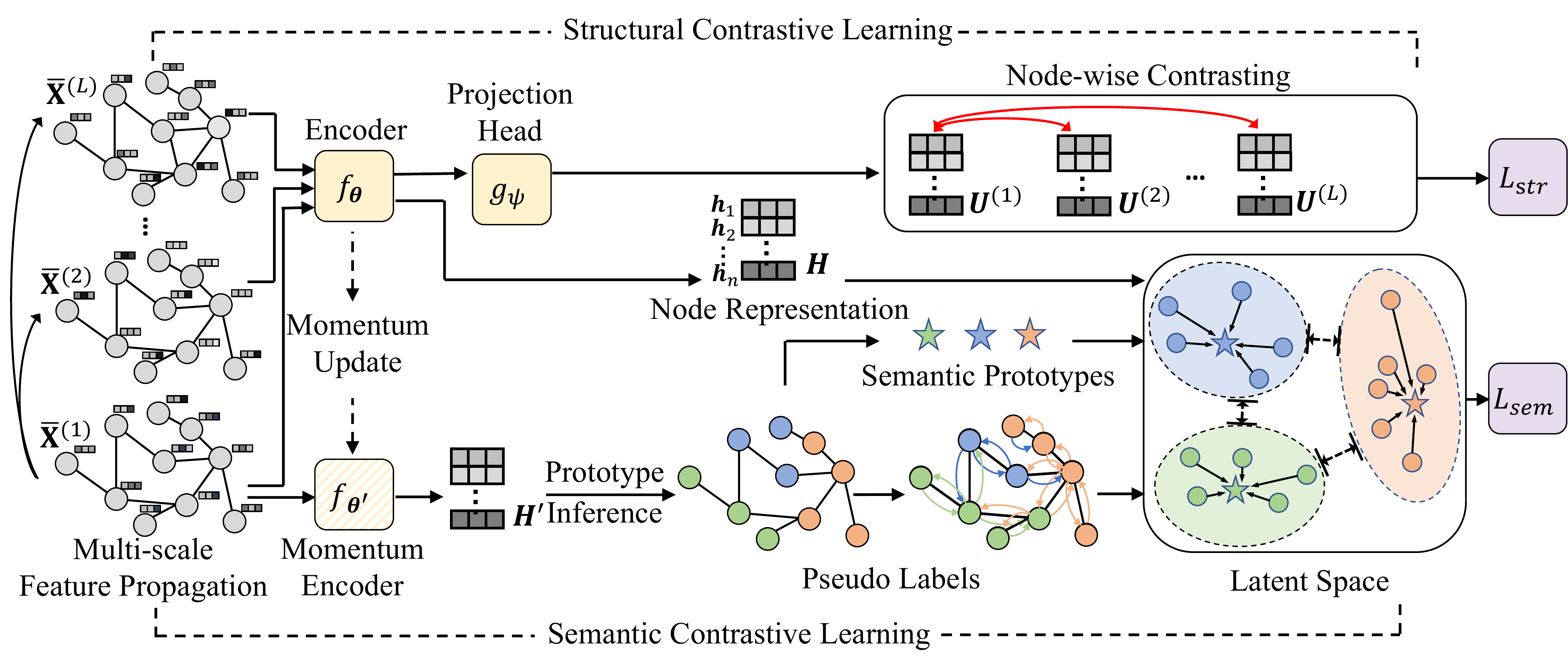}
    }}
    \vspace{-0.2cm}
  \caption{Illustration of the overall framework S$^3$-CL for self-supervised node representation learning.}
  
  \label{fig:overall_framework}
\end{figure*}

In this paper, we propose a novel graph contrastive learning framework S$^3$-CL for unsupervised/self-supervised node representation learning. The overall framework is illustrated in Figure \ref{fig:overall_framework}. Our proposed framework consists of three main components: (i) a simple (e.g., 1-layer) encoder network; (ii) a structural contrastive learning algorithm; and (iii) a semantic contrastive learning algorithm.
    
    


\subsection{Structural Contrastive Learning}

Existing GCL methods for unsupervised node representation learning aim to achieve node-wise discrimination by maximizing the agreement between the representations of the same graph element in different augmented views. Despite their success, they commonly ignore the global structure knowledge due to the limitations of either the adopted data augmentation function or the GNN encoder. In this work, we propose the \textit{structural contrastive learning} algorithm, which enables a simple neural network to capture both local and global structural knowledge by performing contrastive learning on multi-scale augmented graph views.

\noindent\textbf{Multi-scale Feature Propagation.} 
In order to capture long-range node interactions without suffering the oversmoothing issue, in our structural contrastive learning algorithm, we propose to adopt \textit{multi-scale feature propagation} to augment the input graph from the structural perspective. Compared to arbitrarily modifying the graph structure such as perturbing edges or nodes, feature propagation not only allows incorporating long-range node interactions but also mitigates the noises in the original graph~\cite{ding2022data}. Unlike existing GCL algorithms that perform only two augmentations for each instance, we perform feature propagation with different scales to generate $L$ augmented feature matrices $\{\Bar{\mathbf{X}}^{(l)}\}_{l=1}^L$, each of which encodes the $l$-hop node interactions in the graph. Then each augmented feature matrix $\Bar{\mathbf{X}}^{(l)}$ can be encoded by a encoder network $f_{\bm\theta} (\cdot)$ and the corresponding node representations can be computed by:
\begin{equation}
 \mathbf{H}^{(l)} = f_{\bm\theta}(\Bar{\mathbf{X}}^{(l)}) = \text{ReLU} (\Bar{\mathbf{X}}^{(l)} \mathbf{\bm\Theta}),
 \quad \Bar{\mathbf{X}}^{(l)} = \mathbf{T}^{l} \mathbf{X}, 
\label{eq:diff_matrix}
\end{equation}
where $\mathbf{T} \in \mathbb{R}^{N\times N}$ is a generalized transition matrix and we take $\mathbf{T} = \Tilde{\mathbf{A}}_{sym}= {\Tilde{\mathbf{D}}}^{-1/2}\Tilde{\mathbf{A}}\Tilde{\mathbf{D}}^{-1/2}$ in this work. $\mathbf{H}^{(1)}$ is learned from a local view as the message-passing is only enabled between direct neighbors, while $\{\mathbf{H}^{(l)}\}_{l=2}^{L}$ are learned from a set of high-order views that encode the long-range node interactions at different scales.

It is noteworthy that our model inherently separates the feature propagation step, i.e., $\Bar{\mathbf{X}}^{(l)} = \mathbf{T}^{l} \mathbf{X}$, and transformation step, i.e.,
$f_{\bm\theta}(\Bar{\mathbf{X}}^{(l)})$ into the data augmentation and representation learning modules, respectively. Compared to standard GNNs that couple the two steps together in each layer, this decoupling strategy allows the model to perform the high-order feature propagation without conducting non-linear transformations, reducing the risk of over-smoothing~\cite{feng2020graph,ding2022meta} in contrastive learning. In the meantime, we can use a much simpler encoding backbone to transform the augmented features to node representations without stacking multiple GNN layers.

\noindent\textbf{Structural Contrastive Objective.} In our structural contrastive learning, we aim to maximize the agreement between the representations of each node learned from the local view and its different high-order views by maximizing their mutual information. Instead of directly contrasting the output of the encoder network, we follow previous research in contrastive learning~\cite{simclr} and apply a \textit{projection head} $g_{\bm\psi} (\cdot)$ to the node representations computed by the encoder network. As such, the representations we contrast in our structural contrastive learning can be denoted by $\{\mathbf{U}^{(l)}\}_{l=1}^{L}$, where $\mathbf{U}^{(l)} =g_{\bm\psi}( \mathbf{H}^{(l)})$, and $g_{\bm\psi} (\cdot)$ is a two-layer MLP in our implementation.


In our work, we adopt InfoNCE \cite{cpc} to estimate the lower bound of the mutual information between the node representations learned from a local view $\mathbf{U}^{(1)}$ and different high-order views $\{\mathbf{U}^{(l)}\}_{l=2}^L$ of the input graph. The loss function of structural contrastive learning can be defined as:
\begin{equation}
\label{eq:MI_loss}
	\mathcal{L}_{str} = -\sum_{i=1}^N \sum_{l=2}^L \log \frac{\exp(\mathbf{u}^{(1)}_i\cdot \mathbf{u}^{(l)}_i/\tau_1)}{ \sum_{j=1}^{M + L -1} \exp(\mathbf{u}^{(1)}_i\cdot \mathbf{u}^{(l)}_j/\tau_1) } ,
\end{equation}
where $\tau_1$ is the temperature parameter. Note that $\{\mathbf{u}^{(l)}_j\}_{j=1}^{M+L-1}$ contains $L-1$ positive examples and $M$ negative examples sampled from augmented views of other nodes.



By performing the proposed structural contrastive learning based on multi-scale augmentations of the input graph, the encoder network $f_{\bm\theta}(\cdot)$ not only encourages accurate node-wise discrimination but also captures multi-scale global structural knowledge during the learning process. The resulted node representations $\mathbf{H}$ can be computed by feeding the mixed-order propagated features $\Bar{\mathbf{X}}$ to the encoder network as:
\begin{equation}
 \mathbf{H} = f_{\bm\theta}(\Bar{\mathbf{X}}) =  \text{ReLU} (\Bar{\mathbf{X}} \mathbf{\bm\Theta}), \quad \Bar{\mathbf{X}} = \frac{1}{L} \sum_{l= 1}^L \mathbf{T}^{l}\mathbf{X}. 
\label{eq:feature}
\end{equation}
This enables the learned node representations to preserve both local and global structure information compared with directly using $\mathbf{T}^{L}\mathbf{X}$~\cite{xu2018representation,feng2020graph}. 
\subsection{Semantic Contrastive Learning} 
Despite the structural contrastive learning algorithm can provide better node-wise discrimination by exploiting the global structural knowledge based on the multi-scale propagated features, 
it has the same limitation as existing GCL efforts -- cannot explicitly encode the semantic structure of the input graph. To further capture the semantic global knowledge, we propose a \textit{semantic contrastive learning} algorithm that encourages the intra-cluster compactness and inter-cluster separability in the semantic latent space. 

Since the prior knowledge of node clusters is unknown, we propose to iteratively infer the clusters among nodes and the corresponding prototypes based on the learned node representations, and perform semantic contrastive learning to promote those nodes that are semantically similar clustering around their corresponding cluster prototypes. 

We denote the cluster prototype representation via a matrix $\mathbf{C} \in \mathbb{R}^{K \times D'}$, where $K$ is the number of prototypes inferred from the data. We use $\mathbf{c}_k$ to denote the $k$-th row of $\mathbf{C}$, which is the representation of the $k$-th prototype in the latent space. The prototype assignments or pseudo labels of nodes are denoted by $\mathcal{Z} = \{z_i\}_{i=1}^n$, where $z_i\in \{1,...,K\}$ is the pseudo label of node $v_i$.

\noindent\textbf{Bayesian Non-parametric Prototype Inference.} A key function of our semantic contrastive learning algorithm is to infer highly representative cluster prototypes. However, the optimal number of clusters is unknown under the setting of unsupervised node representation learning. To bridge the gap, we propose a Bayesian non-parametric prototype inference algorithm to approximate the optimal number of clusters and simultaneously compute the cluster prototypes. Specifically, we build a Dirichlet Process Mixture Model (DPMM) and assume the distribution of node representations is a mixture of Gaussians, in which each component is used to model the prototype of a cluster. Note that the components share the same fixed covariance matrix $\sigma \mathbf{I}$. The DPMM model is defined as:
\begin{equation}
\begin{aligned}
G  \sim  \text{DP} (G_0, \alpha), \quad \bm\phi_i \sim  G , \quad \mathbf{h}_i \sim  \mathcal{N}(\bm\phi_i,\sigma  \mathbf{I}),
\end{aligned}
\label{eq:DPMM}
\end{equation}
where $G$ is a Gaussian distribution drawn from the Dirichlet process $\mbox{DP}(G_0, \alpha)$, and $\alpha$ is the concentration parameter for $\mbox{DP}(G_0, \alpha)$. $\bm \phi_i$ is the mean of the Gaussian sampled for node representation $\mathbf{h}_i$. $G_0$ is the prior over means of the Gaussians. We take $G_0$ to be a zero-mean Gaussian ${\mathcal N}(\bm{0},\rho \mathbf{I})$, where $\rho \mathbf{I}$ is the covariance matrix.

Next, we use a collapsed Gibbs sampler \cite{resnik2010gibbs} to infer the Gaussian components. The Gibbs sampler iteratively samples pseudo labels for the nodes given the means of the Gaussian components and samples the means of the Gaussian components given the pseudo labels of the nodes. Following \cite{kulis2011revisiting}, such a process is almost equivalent to K-Means when the variance of the Gaussian components $\sigma\rightarrow 0$. The almost zero variance eliminates the need to estimate the variance $\sigma$, thus making the inference efficient. Let $\Tilde{K}$ denote the number of inferred prototypes at the current iteration step, the prototype assignment update can be formulated as:
\begin{equation}
\label{eq:label_update}
\begin{aligned}
  z_{i} &= \argmin_{k} \left\{ d_{ik} \right\},\\
  d_{ik} &=  
          \left\{\begin{array}{l l}
          ||\mathbf{h}_{i} - \mathbf{c}_{k}||^2  & \text{for~} k = 1,..., \Tilde{K} \\
          \xi  & \text{for~}k = \Tilde{K}+1,
        \end{array}\right.
\end{aligned}
\end{equation} 
where $d_{ik}$ is the distance to determine the pseudo labels of node representation $\mathbf{h}_i$. $\xi$ is the margin to initialize a new prototype. 
In practice, we choose the value of $\xi$ by performing cross-validation on each dataset.
With the formulation in Equation (\ref{eq:label_update}), a node will be assigned to the prototype modeled by the Gaussian component corresponding to the closest mean of Gaussian, unless the squared Euclidean distance to the closest mean is greater than $\xi$. 
After obtaining the pseudo labels, the cluster prototype representations can be computed by:
    $\mathbf{c}_k ={\sum_{z_i = k}\mathbf{h}_i}/{\sum_{z_i = k}1}$, for $k=1,...,\Tilde{K}$.

Note that we iteratively update prototype assignments and prototype representations till convergence, and we set the number of prototypes $K$ to be the number of inferred prototypes $\Tilde{K}$ afterward. Afterwards, we refine the cluster prototypes using label propagation and we attach the details in Supplementary \ref{sec:label} due to the space limit.

\noindent\textbf{Semantic Contrastive Objective.}
After obtaining the prototype assignments $\mathcal{Z}$ and prototype representations $\mathbf{C}$, our semantic contrastive objective aims to consolidate the semantic structure (i.e., intra-cluster compactness and inter-cluster separability) of the learned node representation $\mathbf{H}$ by updating the encoder parameter $\bm\theta$. To this end, we maximize the likelihood of each node in the graph given $\bm\theta$ and $\mathbf{C}$:
\begin{equation}
\begin{split}
\label{eq:log-likelihood}
    Q(\bm\theta) &=  \sum_{n=1}^{N} \log p(\mathbf{x}_i| \bm\theta, \mathbf{C}) \\
    &=  \sum_{n=1}^{N} \log \sum_{k=1}^{K} p(\mathbf{x}_i, k| \bm\theta, \mathbf{C}),
\end{split}
\end{equation}
where $p$ is the probability density function. Directly optimizing log-likelihood $Q(\bm\theta)$ is intractable as the labels of nodes are unknown. Instead, we optimize the variational lower bound of $Q(\bm\theta)$, given by:
\begin{equation}
\label{eq:VLB}
\begin{split}
Q(\bm\theta)\geq&\sum_{i=1}^N \sum_{k=1}^K p(k|\mathbf{x}_i)\log \frac{p(\mathbf{x}_i, k|\bm\theta, \mathbf{C})}{p(k|\mathbf{x}_i)}\\
 = &\sum_{i=1}^N\sum_{k=1}^K p(k|\mathbf{x}_i)\log p(\mathbf{x}_i, k|\bm\theta, \mathbf{C})\\
  &- \sum_{i=1}^N\sum_{k=1}^K p(k|\mathbf{x}_i)\log p(k|\mathbf{x}_i).
\end{split}
\end{equation}
Note that we can drop the second term of the right-hand side of Equation~(\ref{eq:VLB}) as it is a constant. To maximize the remaining part $\sum_{i=1}^N\sum_{k=1}^K p(k|\mathbf{x}_i)\log p(\mathbf{x}_i, k|\bm\theta, \mathbf{C})$, we can estimate $p(k|\mathbf{x}_i)$ by $p(k|\mathbf{x}_i, \bm\theta, \mathbf{C}) = \mathbb{1}_{\{k=z_i\}}$, as we assign $\mathbf{x}_i$ to cluster $z_i$ given $\mathbf{C}$ in our DPMM model. Thus, we can maximize $Q(\bm\theta)$ by minimizing the following loss function:
\begin{equation}
\label{eq:bound}
 \mathcal{L}_{sem} = -\sum_{i=1}^N\log p(\mathbf{x}_i, z_i|\bm\theta, \mathbf{C}).
\end{equation}

Under the assumption of a uniform prior distribution of $\mathbf{x}_i$ over different prototypes, we have $p(\mathbf{x}_i, z_i| \bm\theta, \mathbf{C}) \propto p(\mathbf{x}_i| z_i,  \bm\theta, \mathbf{C})$. 
Since the distribution of node representation around each prototype generated by the DPMM is an isotropic Gaussian, after applying $\ell_2$ normalization on the representation of nodes and prototypes, 
we can estimate $p(\mathbf{x}_i| z_i,  \bm\theta, \mathbf{C})$ by: 
\begin{equation}
\label{eq:p_xi}
    p(\mathbf{x}_i| z_i,  \bm\theta, \mathbf{C}) = \frac{\exp(\mathbf{h}_i\cdot \mathbf{c}_{z_i}/\tau_2)}{\sum_{k=1}^K \exp(\mathbf{h}_i \cdot \mathbf{c}_k/\tau_2)},
\end{equation}
where $\mathbf{c}_{z_i}$ is the representations of $z_i$-th prototype. The temperature parameter $\tau_2 \propto \sigma^2$ is related to the concentration of node representation around each prototype, and $\sigma$ is the variance of the Gaussians in the DPMM model defined by Equation~(\ref{eq:DPMM}). For the simplicity of training, we directly take $\tau_2$ as a hyperparameter. Taking Equation~(\ref{eq:p_xi}) into Equation~(\ref{eq:bound}), we can maximize $Q(\bm\theta)$ by minimizing the following loss function:
\begin{equation}
\label{eq:gaussian_lower}
    \mathcal{L}_{sem} = -\sum_{i=1}^N\log\frac{\exp(\mathbf{h}_i\cdot \mathbf{c}_{z_i}/\tau_2)}{\sum_{k=1}^K \exp(\mathbf{h}_i \cdot \mathbf{c}_k/\tau_2)}.
\end{equation}

\subsection{Model Learning} 
Given the proposed S$^3$-CL learning framework, our goal is to learn expressive node representations that preserve both valuable structural and semantic knowledge without any semantic labels. In this section, we will introduce the overall loss function, and also the optimization of the proposed framework with regard to the network parameters, prototype assignments, and prototype representations. 

\noindent\textbf{Overall Loss.} To train our model in an end-to-end fashion and learn the encoder $f_{\bm\theta}(\cdot)$, we jointly optimize both the structural and semantic contrastive learning losses. The overall objective function is defined as:
\begin{equation}
\label{eq:overall_loss}
    \mathcal{L} = \gamma \mathcal{L}_{str} + (1-\gamma) \mathcal{L}_{sem},
\end{equation}
where we aim to minimize $\mathcal{L}$ during training, and $\gamma$ is a balancing parameter to control the contribution of each contrastive learning loss. For the sake of the stability of the training of the encoder, we apply our Bayesian non-parametric prototype inference algorithm on the node representations computed by a momentum encoder~\cite{moco}. 

Notably, in semantic contrastive learning, the computed pseudo labels $\mathcal{Z}$ can be utilized in the negative example sampling process in our structural contrastive learning to avoid sampling bias issues \cite{chuang2020debiased}. We select negative samples in Equation~(\ref{eq:MI_loss}) for each node from nodes assigned to different prototypes. 


\begin{algorithm}[t!]
\caption{The learning algorithm of S$^3$-CL.}
\label{alg:Algorithm-DGCL}
\LinesNumbered
\small
\KwIn{Attribute matrix $\mathbf{X}$; adjacency matrix $\mathbf{A}$; propagation step $L$}
\KwOut{Pretrained encoder network $f_{\bm\theta} (\cdot)$}

Initialize encoder parameter $\bm\theta$ and $\bm\theta'$

\While{not converge}{
    
Compute node representations of different augmented views $\{\mathbf{H}^{(l)}\}_{l=1}^{L}$ and $\{\mathbf{U}^{(l)}\}_{l=1}^{L}$

Compute the prototype representations $\mathbf{C}$ and prototype assignments $\mathcal{Z}$ \Comment{E-step Update}


Calculate loss $\mathcal{L}_{str}$ and $\mathcal{L}_{sem}$ by Equation (\ref{eq:MI_loss}) and Equation (\ref{eq:gaussian_lower}), respectively

$\mathcal{L} = \gamma \mathcal{L}_{str} + (1-\gamma) \mathcal{L}_{sem}$

Update $\bm\theta$ by minimizing $\mathcal{L}$ \Comment{M-step Update}

Update momentum encoder $\bm\theta'$
}

\Return the encoder network $f_{\bm\theta} (\cdot)$
\end{algorithm}

\noindent\textbf{Model Optimization via EM.} Specifically, we adopt EM algorithm to alternately estimate the posterior distribution $p(z_i|\mathbf{x}_i,\bm\theta, \mathbf{C})$ and optimize the network parameters $\bm\theta$. We describe the details for the E-step and M-step applied in our methods as follows:
\begin{itemize}[leftmargin=*,noitemsep,topsep=1.5pt]
    \item \textbf{E-step.} In this step, 
    we fix the network parameter $\bm\theta$, and estimate the prototypes $\mathbf{C}$ and the prototype assignment $\mathcal{Z}$ with our proposed Bayesian non-parametric prototype inference algorithm (more details in Supplementary \ref{sec:Algorithm-Proto}). 

    \item \textbf{M-step.} Given the posterior distribution computed by the E-step, we aim to maximize the expectation of log-likelihood $Q(\bm\theta)$, by directly optimizing the semantic contrastive loss function $\mathcal{L}_{sem}$. In order to perform structural and semantic contrastive learning at the same time, we instead optimize a joint overall loss function as formulated in Equation (\ref{eq:overall_loss}).
\end{itemize}

Algorithm \ref{alg:Algorithm-DGCL} outlines the learning process of the proposed framework. After the self-supervised pre-training is done, the pre-trained encoder can be directly used to generate node representations for various downstream tasks. 
\section{Experiments}
\subsection{Experimental Settings}

\noindent\textbf{Evaluation Datasets.} In our experiments, we evaluate S$^3$-CL on six public benchmark datasets that are widely used for node representation learning, including Cora \cite{sen_2008_aimag}, Citeseer \cite{sen_2008_aimag}, Pubmed \cite{namata2012query}, Amazon-P \cite{shchur2018pitfalls}, Coauthor CS \cite{shchur2018pitfalls} and ogbn-arxiv \cite{hu2020open}. Cora, Citeseer, and Pubmed are the three most widely used citation networks.
Amazon-P is a co-purchase graph and Coauthor CS is a co-authorship graph. 
The ogbn-arxiv is a large-scale citation graph benchmark dataset.


\noindent\textbf{Compared Methods.}
To demonstrate the effectiveness of our proposed method, six state-of-the-art graph self-supervised learning methods are compared in our experiments, including DGI~\cite{velickovic_2019_iclr}, MVGRL~\cite{hassani2020contrastive}, GMI~\cite{gmi}, GRACE~\cite{Zhu:2020vf}, MERIT~\cite{jin2021multi}, and SUGRL~\cite{mo2022simple}. As we consider node classification as our downstream task, we also include five representative supervised node classification methods, namely MLP~\cite{velickovic_2019_iclr}, LP~\cite{zhu2003semi}, GCN~\cite{kipf2017semi}, GAT~\cite{gat}, and SGC~\cite{sgc}, as baselines for the evaluation on the node classification task. To evaluate the model performance for node clustering, we compare S$^3$-CL against methods including K-Means~\cite{lloyd1982least}, GAE \cite{kipf_2016_arxiv}, adversarially regularized GAE (ARGA) and VGAE (ARVGA)~\cite{pan_2018_ijcai}, GALA \cite{park_2019_iccv}, DGI, DBGAN~\cite{zheng2020distribution}, MVGRL, MERIT, and SUGRL.

\subsection{Evaluation Results}
\begin{table*}[t]

\centering
\scalebox{0.9}{
\begin{tabular}{lcccccc} 
\toprule
Methods      & Cora            & Citeseer        & Pubmed     &Amazon-P     & Coauthor CS      & ogbn-arxiv  \\ 
\hline
\multicolumn{7}{c}{\textsc{supervised}}                                                                           \\ 
\hline
MLP & $55.2\pm{0.4}$ & $46.5\pm{0.5}$ & $71.4\pm{0.3}$  									& $78.5\pm{0.2}$	& $76.5\pm{0.3}$ & $55.5\pm{0.2}$\\
LP \cite{zhu2003semi}  & $68.0\pm{0.5}$      & $45.3\pm{0.6}$            & $63.0\pm{0.3}$ 	& $75.4\pm{0.0}$	& $74.3\pm{0.0}$ &   $68.3\pm{0.0}$                 \\
GCN~\cite{kipf2017semi}  & $81.7\pm{0.4}$ & $70.5\pm{0.3}$ & $79.4\pm{0.4}$  				& $87.3\pm{1.0}$	& $91.8\pm{0.1}$ & $71.7\pm{0.3}$  \\
GAT~\cite{gat}  & $83.0\pm{0.7}$ & $72.5\pm{0.7}$ & $79.0\pm{0.3}$ 							& $86.2\pm{1.5}$	& $90.5\pm{0.7}$ &   $\textbf{73.2}\pm\textbf{0.2}$                 \\
SGC~\cite{sgc}         & $81.5\pm{0.2}$ & $73.1\pm{0.1}$ & $79.7\pm{0.4}$ 					& $88.3\pm{1.1}$	& $91.5\pm{0.3}$ &       $69.8\pm{0.2}$             \\ 
\hline
\multicolumn{7}{c}{\textsc{self-supervised} + \textsc{fine-tuning}}                               \\ 
\hline
DGI~\cite{velickovic_2019_iclr} & $81.7\pm{0.6}$ & $71.5\pm{0.7}$ & $77.3\pm{0.6}$ 				& $83.1\pm{0.3}$	& $90.0\pm{0.3}$ & $67.1\pm{0.4}$  \\
GMI~\cite{gmi}         & $82.7\pm{0.2}$ & $73.0\pm{0.3}$ & $80.1\pm{0.2}$			& $85.1\pm{0.0}$	& $91.0\pm{0.0}$ & $69.6\pm{0.3}$                   \\
MVGRL~\cite{hassani2020contrastive}       & $82.9\pm{0.7}$ & $72.6\pm{0.7}$ & $79.4\pm{0.3}$ 	& $87.3\pm{0.1}$	& $91.3\pm{0.1}$ &  $71.3\pm{0.2}$ \\
GRACE~\cite{Zhu:2020vf}       & $80.0\pm{0.4}$ & $71.7\pm{0.6}$ & $79.5\pm{1.1}$ 					& $81.8\pm{0.8}$	& $90.1\pm{0.8}$ & $71.1\pm{0.2}$\\
MERIT~\cite{jin2021multi}       & $83.1\pm{0.6}$ & \underline{$74.0\pm{0.7}$} & $80.1\pm{0.4}$ 	& \underline{$88.8\pm0.4$}					& \underline{$92.4\pm{0.4}$} &$71.7\pm{0.1}$   \\
SUGRL~\cite{mo2022simple}       & \underline{$83.4\pm{0.5}$} & $73.0\pm{0.5}$ & $\textbf{81.9}\pm\textbf{0.5}$ 	& $88.5\pm{0.2}$					& $92.2\pm{0.5}$ & $69.3\pm0.2$   \\
S$^3$-CL (ours) & $\textbf{84.5}\pm\textbf{0.4}$ & $\textbf{74.6}\pm\textbf{0.4}$ & \underline{$80.8\pm0.3$ }		& $\textbf{89.0}\pm\textbf{0.5}$					& $\textbf{93.1}\pm\textbf{0.4}$ & \underline{$72.8\pm0.3$}                   \\
\bottomrule
\end{tabular}}
\vspace{-0.1cm}
\caption{Node classification performance comparison on benchmark datasets.}
\label{tab:classification}
\end{table*}

\begin{table*}[t]
\centering
\scalebox{0.9}{
\begin{tabular}{lccccccccccc} 
\toprule
\multirow{2}{*}{Methods} & \multicolumn{3}{c}{Cora}                    &  & \multicolumn{3}{c}{Citeseer}                 & & \multicolumn{3}{c}{Pubmed}                     \\ 
\cline{2-4} \cline{6-8} \cline{10-12}
                         & ACC           & NMI           & ARI        &   & ACC           & NMI           & ARI         &  & ACC           & NMI           & ARI            \\ 
\hline
K-Means                & 49.2          & 32.1          & 22.9          && 54.0          & 30.5          & 27.8          && 59.5          & 31.5          & 28.1           \\
GAE~\cite{kipf_2016_arxiv}                      & 59.6          & 42.9          & 34.7          && 40.8          & 17.6          & 12.4          && 67.2          & 27.7          & 27.9           \\
ARGA~\cite{pan_2018_ijcai}                      & 64.0          & 44.9          & 35.2          && 57.3          & 35.0          & 34.1          && 66.8          & 30.5          & 29.5           \\
ARVGA~\cite{pan_2018_ijcai}                    & 64.0          & 45.0          & 37.4          && 54.4          & 26.1          & 24.5          && 69.0          & 29.0          & 30.6           \\
GALA~\cite{park_2019_iccv}                    & 74.5          & 57.6        & 53.1          && 69.3          & 44.1          &44.6      && 69.3          & 32.7          & 32.1           \\
DGI~\cite{velickovic_2019_iclr}                     & 55.4          & 41.1          & 32.7          && 51.4          & 31.5          & 32.6          && 58.9          & 27.7          & 31.5            \\
DBGAN~\cite{zheng2020distribution}                  & \underline{74.8}          & 56.0          & \underline{54.0}          && 67.0          & 40.7          & 41.4          && 69.4          & 32.4          & 32.7           \\
MVGRL~\cite{hassani2020contrastive}                   & 73.2          & 56.2          & 51.9          && 68.1          & 43.2          & 43.4          && 69.3          & 34.4          & 32.3           \\
MERIT~\cite{jin2021multi}                    & 73.6          & 57.1          & 52.8         && 68.9          & 43.9          & 44.1          && \underline{69.5}          & 34.7          & 32.8        \\
SUGRL~\cite{mo2022simple}                    & 73.9          & \underline{58.5}          & 53.0        && \underline{70.5}          & \underline{45.8}          & \underline{47.0}          && \underline{69.5}         & \underline{35.0}        & \underline{33.4}           \\
S$^3$-CL (ours)                     & \textbf{75.1} & \textbf{60.7} & \textbf{56.6} && \textbf{71.2} & \textbf{46.3} & \textbf{48.5} && \textbf{71.3} & \textbf{36.0} & \textbf{34.7}  \\
\hline
\end{tabular} 
}
\vspace{-0.1cm}
\caption{Node clustering performance comparison on benchmark datasets.}
\label{tab:clustering}

\end{table*}

\noindent\textbf{Node Classification.}
To evaluate the trained encoder network, we adopt a linear evaluation protocol by training a separate logistic regression classifier on top of the learned node representations. We follow the evaluation protocols in previous works~\cite{velickovic_2019_iclr,hu2020open} for node classification. The mean classification accuracy with standard deviation on the test nodes after 10 runs of training is reported. To avoid the out-of-memory issue when evaluating MVGRL, GRACE, and MERIT on the ogbn-arxiv dataset, we subsample 512 nodes as negative samples for each node during the self-supervised learning phase.

The node classification results of different methods are reported in Table \ref{tab:classification}. We can clearly see that S$^3$-CL outperforms the state-of-the-art self-supervised node representation learning methods across the five public benchmarks. 
Such superiority mainly stems from two factors: (i) our approach S$^3$-CL grants each node access to information of nodes in a larger neighborhood; (ii) S$^3$-CL infers the semantic information of nodes, and enforces intra-cluster compactness and inter-cluster separability on the node representation. With the help of this extra information, node representations generated by S$^3$-CL are more informative and distinctive. Without access to labels, S$^3$-CL even outperforms supervised methods like SGC and GAT.
\begin{figure*}[t]
    \centering
    \scalebox{1.0}{

    \subfigure[Cora]
    {
    \includegraphics[height=0.18\textwidth]{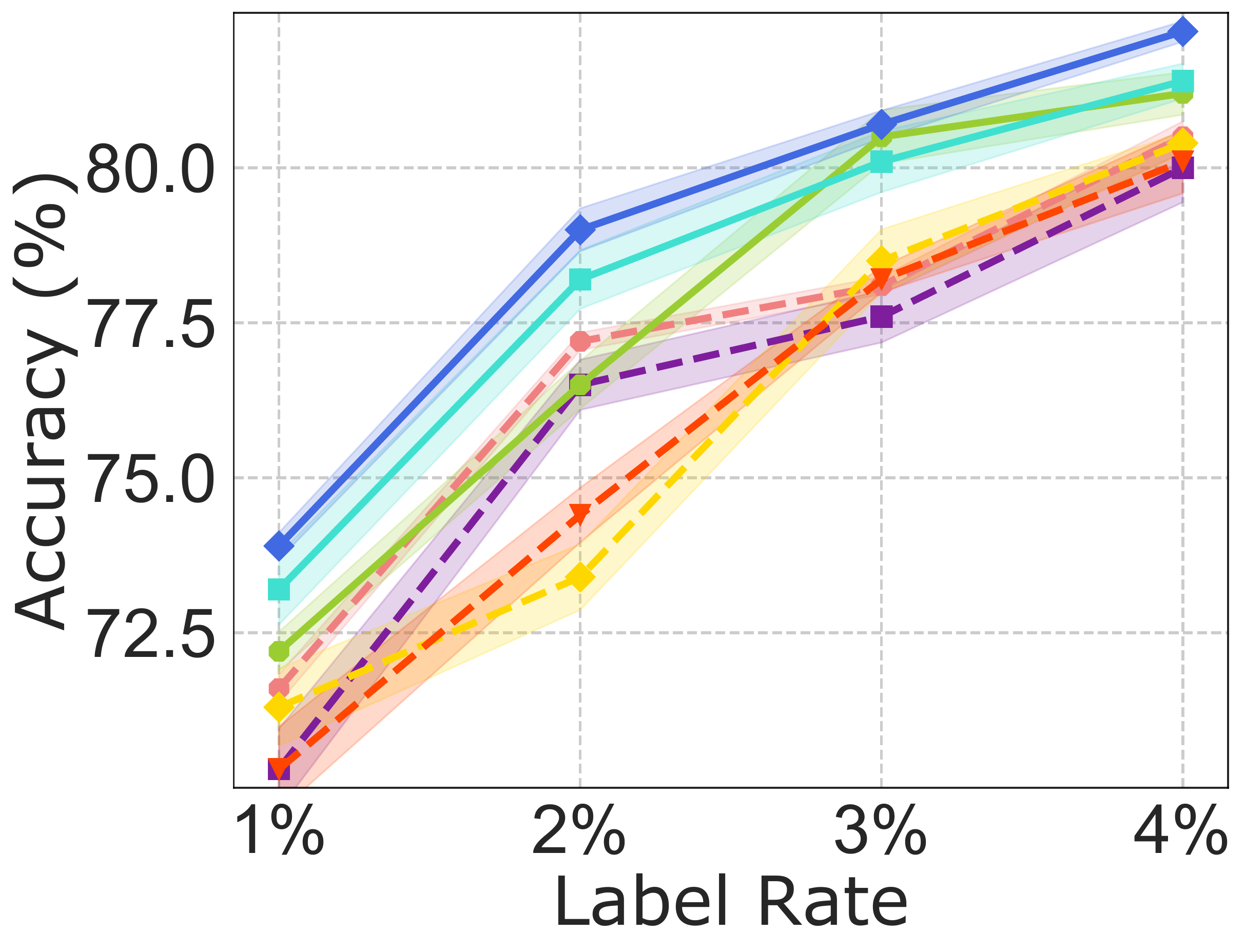}
    }
   
    \subfigure[Citeseer]
    {
    \includegraphics[height=0.18\textwidth]{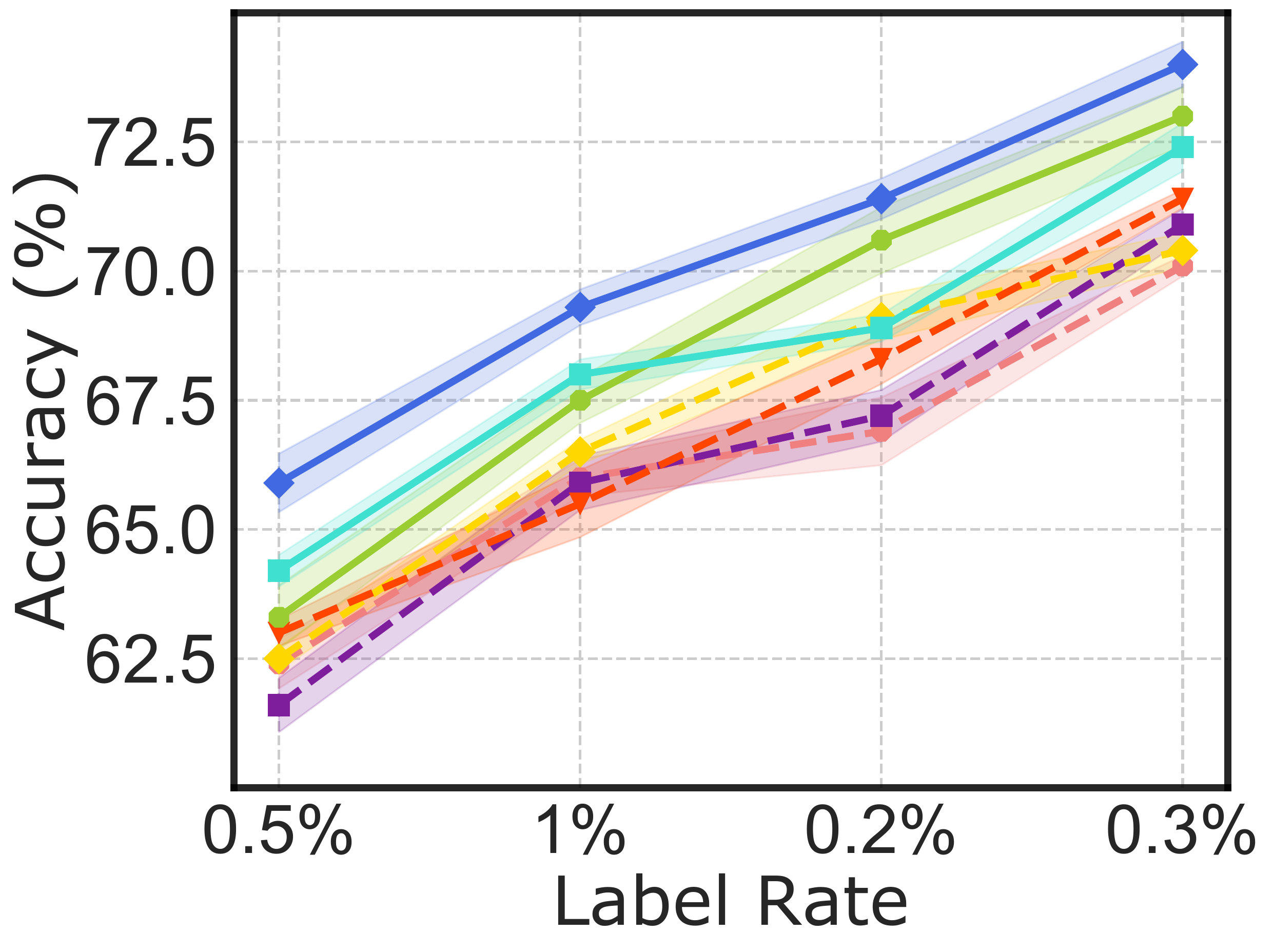}
    }
  
    \subfigure[Pubmed]
    {
    \includegraphics[height=0.18\textwidth]{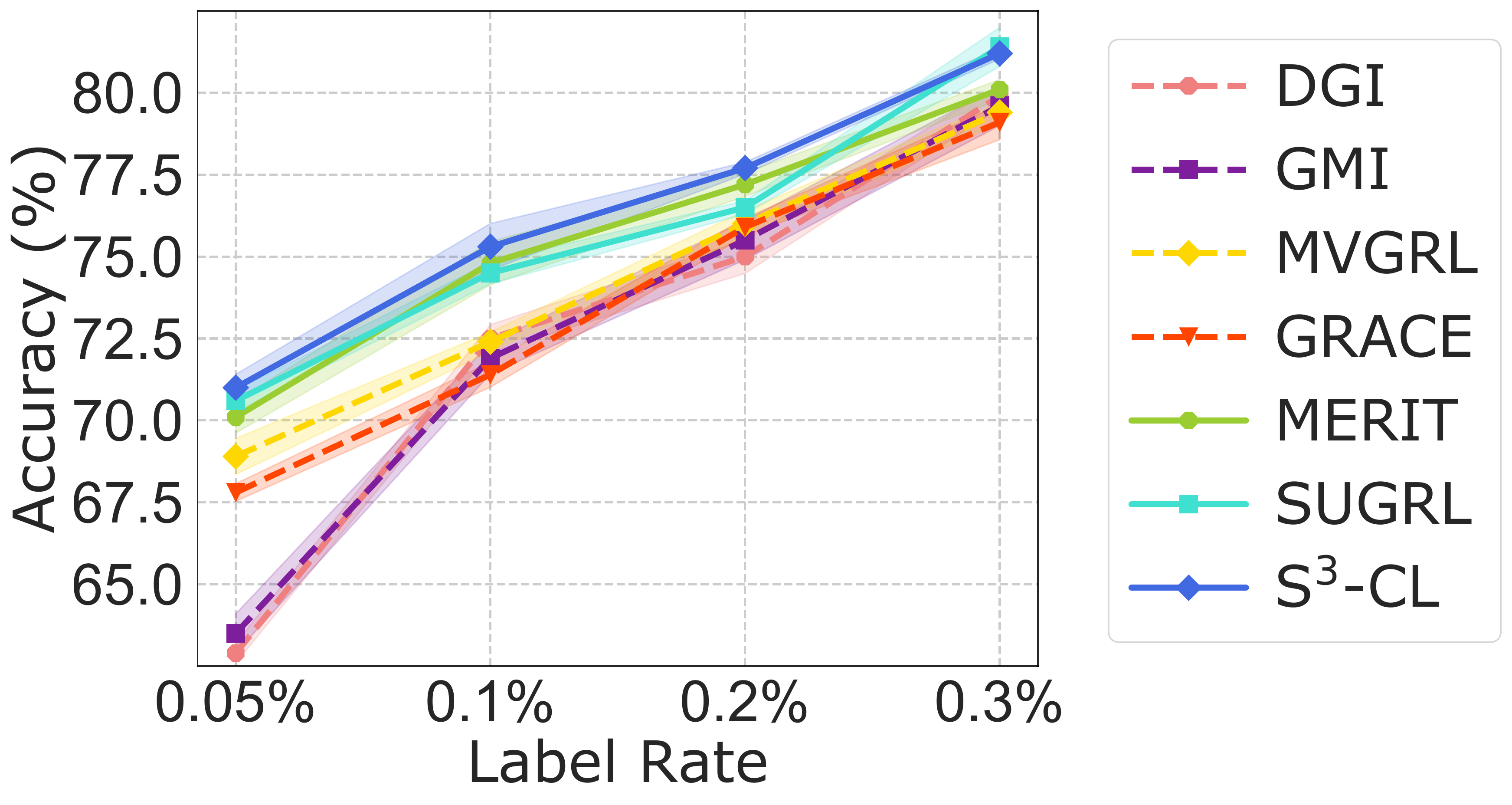}
    }
    }
    \vspace{-0.25cm}
    \caption{Node classification results with limited training labels.}%
    
    \label{fig:label_rate}

\end{figure*}
\begin{figure*}[t]
    \centering
    \scalebox{1.0}{

    \subfigure[Cora]
    {
    \includegraphics[height=0.18\textwidth]{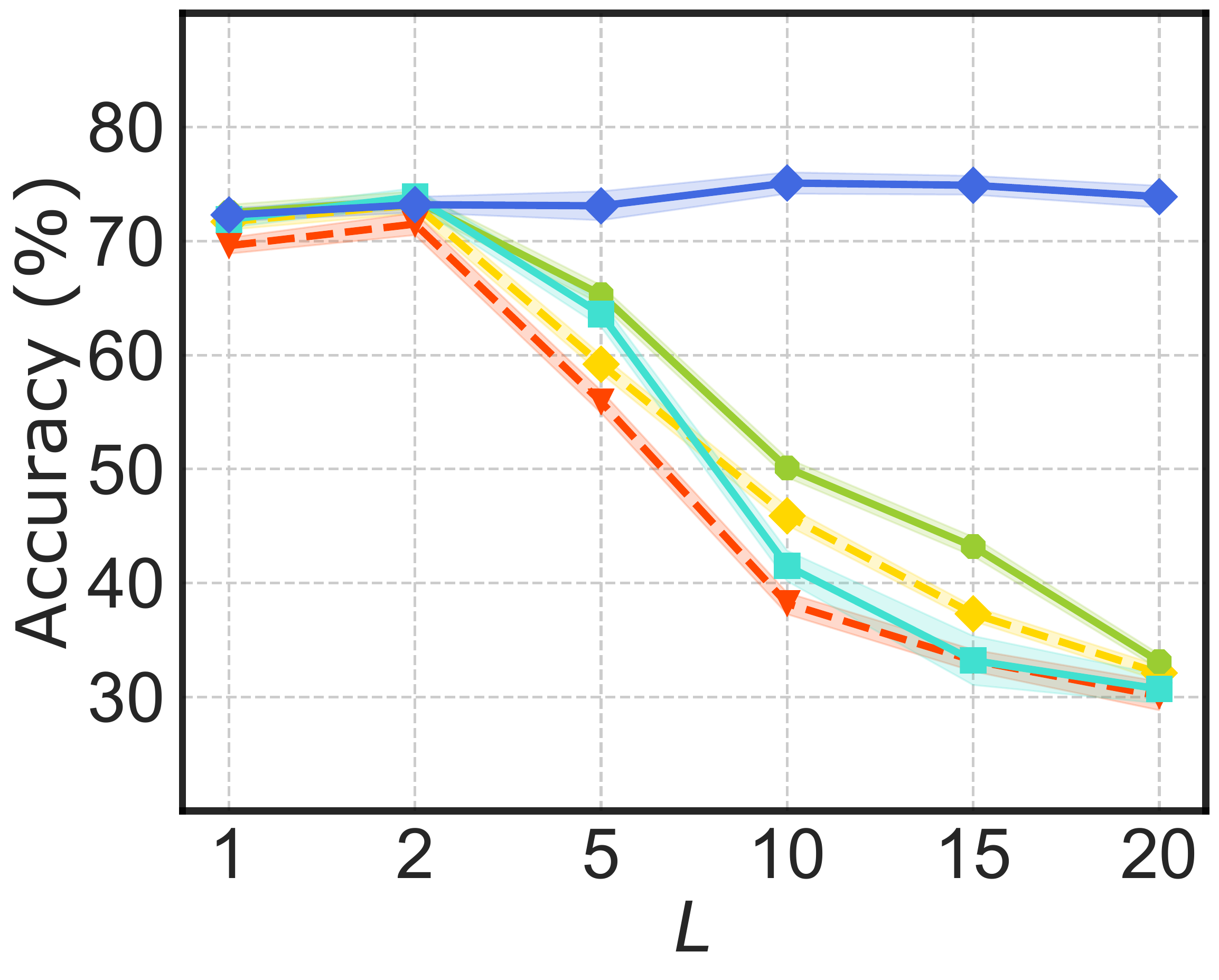}
    }
   
    \subfigure[Citeseer]
    {
    \includegraphics[height=0.18\textwidth]{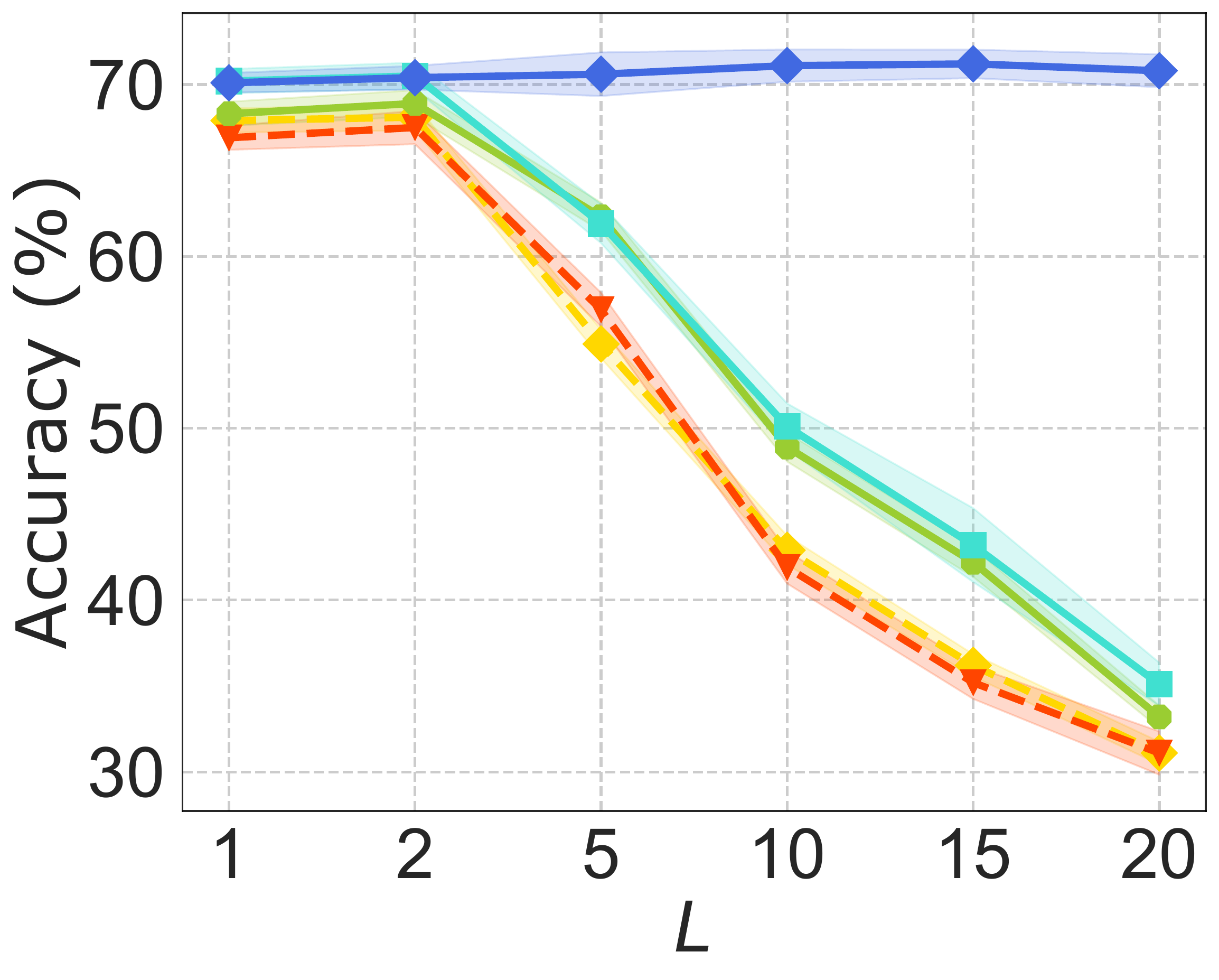}
    }
  
    \subfigure[Pubmed]
    {
    \includegraphics[height=0.18\textwidth]{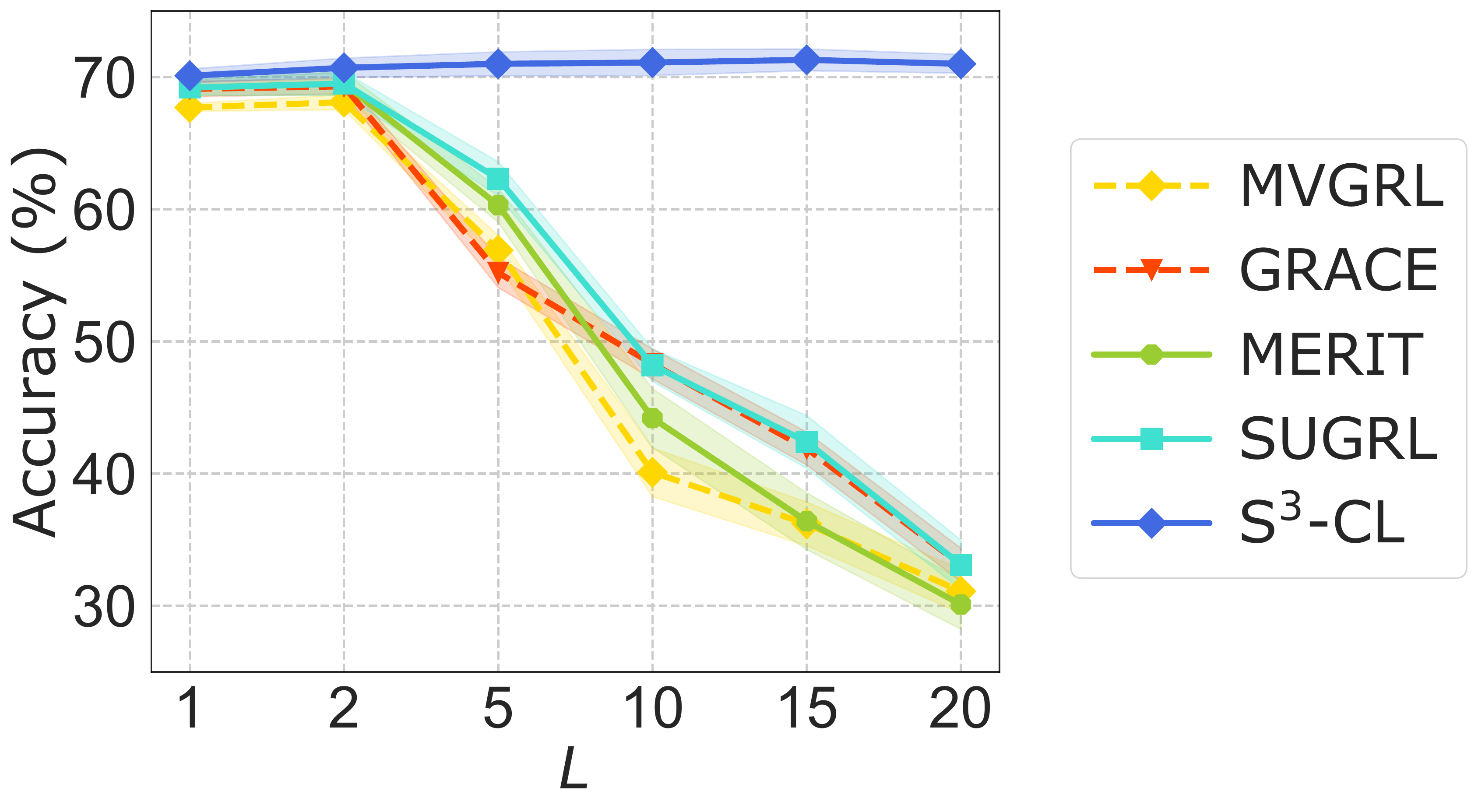}
    }
    }
    \vspace{-0.25cm}
    \caption{Node clustering results of GCL methods with various propagation steps ($L$).}%
    \label{fig:aggregation}
\end{figure*}

\noindent\textbf{Node Clustering.}
To evaluate the quality of the node representations learned by different methods, we conduct experiments on node clustering. We follow the same evaluation protocol as in \cite{hassani2020contrastive}. K-Means is applied on the learned node representation to get clustering results. We use accuracy (ACC), normalized mutual information (NMI), and adjusted rand index (ARI) to measure the performance of clustering. We report the averaged clustering results over 10 times of execution. 

The clustering results are displayed in Table~\ref{tab:clustering}. It is observed that our approach achieves remarkable performance gain over compared methods. For example, the NMI on Cora is improved by $2.2\%$ against the previous SOTA method SUGRL. Such improvement greatly attributes to the fact that S$^3$-CL explores the semantic information of nodes instead of enforcing node-wise discrimination alone as other GCL methods. Thus, the node representation learned by S$^3$-CL works well for clustering algorithms.

\noindent\textbf{Node Classification with Few Labels.} We further evaluate the impact of label rate on the downstream node classification task. Specifically, we evaluate all self-supervised learning methods from Table \ref{tab:classification} under different low-resource settings. The results in Figure~\ref{fig:label_rate} show that our proposed framework S$^3$-CL can still outperform existing methods given a lower label rate. It validates that the node representations learned by our approach S$^3$-CL can encode valuable structural and semantic knowledge from the input graph. As a result, the node representations can be effectively used for the node classification task even with an extremely small label ratio. 

\noindent\textbf{Effect of Feature Propagation.} Next, we investigate the effect of multi-scale feature propagation in the structural contrastive learning by altering the propagation steps $L$. A larger $L$ allows message-passing within a larger neighborhood for learning the node representations. To demonstrate the power of our approach in utilizing structural global knowledge, we compare S$^3$-CL against GRACE, MVGRL, MERIT, and SUGRL with different numbers of layers $L$. The node clustering accuracy of different methods is shown in Figure~\ref{fig:aggregation}. By increasing the propagation steps (number of layers), we can clearly observe that existing unsupervised GCL methods severely degrade due to the oversmoothing issue. In contrast, S$^3$-CL consistently achieves improved performance by making use of information in a larger neighborhood for node representation learning.

\noindent\textbf{Ablation Study.} To validate the effectiveness of the structural contrastive learning and semantic contrastive learning in S$^3$-CL, we conduct an ablation study on Citesser, Cora, and Pubmed with two variants of S$^3$-CL, each of which has one of the contrastive learning components removed. The node classification results are shown in Table~\ref{tab:ablation}. We can observe that the performance of S$^3$-CL degrades when any of the components are removed. Our S$^3$-CL using all components achieves the best performance as the structural and semantic contrastive components complement each other. Hence, the effectiveness of each component is verified.

\begin{table}[!htbp]
	\small
	\centering
	\begin{tabular}{lccc}
		\toprule
		Method & Citeseer & Cora&Pubmed\\
		\midrule
		w/o structural & 73.1$\pm{0.2}$&83.3$\pm{0.3}$  & 80.0$\pm$ 0.3\\
		w/o semantic & 71.9$\pm{0.4}$& 82.2$\pm{0.5}$ &79.3$\pm$ 0.2\\
		S$^3$-CL&  \textbf{74.6}$\pm{\textbf{0.4}}$ &\textbf{84.5}$\pm{\textbf{0.4}}$&\textbf{80.8}$\pm{\textbf{0.3}}$ \\
		\bottomrule
	\end{tabular}
	\vspace{-0.2cm}
	\caption{Ablation study on contrastive components.}
	\label{tab:ablation}
\end{table}

\begin{figure}[h]
    \subfigure[MERIT]{
        \includegraphics[width=0.45\columnwidth]{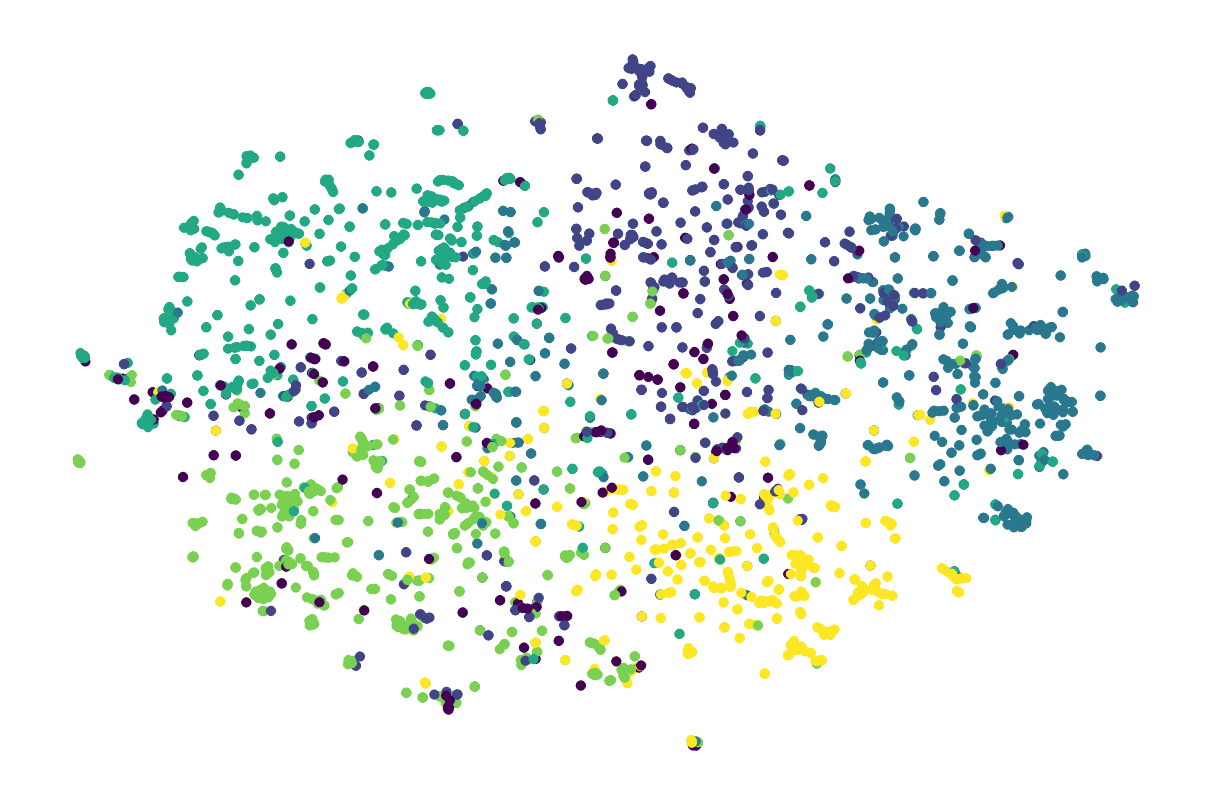}
    }
    \hspace{-0.2cm}
     \subfigure[S$^3$-CL]{
        \includegraphics[width=0.45\columnwidth]{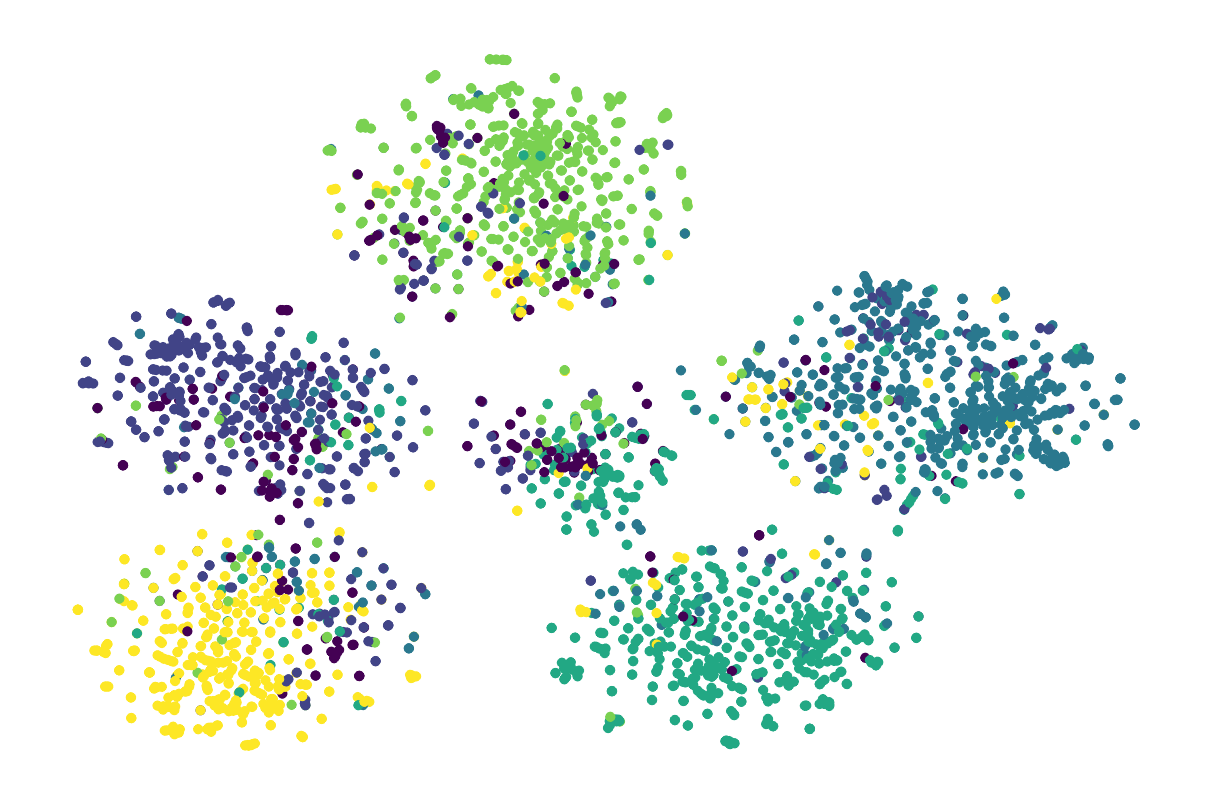}
    }
    \caption{Representation visualization on the Citeseer dataset.}
    \vspace{-0.2cm}
     \label{fig:tsne}
\end{figure}

\noindent\textbf{Representation Visualization.}
To visually show the superior quality of the node representations learned by S$^3$-CL, we use t-SNE to visualize and compare the learned node representations between S$^3$-CL and the best-performing baseline on Citeseer, i.e., MERIT. The visualization results are shown in Figure~\ref{fig:tsne}, where each dot represents the representation of a node, and the color of the dot denotes its ground-truth label. From the figure, we can observe that though some classes can be identified by MERIT, the boundaries between different classes are unclear. Our proposed model is able to enforce better intra-cluster compactness and inter-cluster separability.

\vspace{-0.2cm}
\section{Conclusion}In this paper, we propose a new GCL framework named S$^3$-CL, which can effectively capture the global knowledge from both structural and semantic perspectives for unsupervised node representation learning. By jointly optimizing the structural and semantic contrastive learning losses, we can build the encoder network with simple neural networks to learn expressive node representations for different downstream tasks without using any human-annotated labels. We conduct extensive experiments and demonstrate that S$^3$-CL can outperform the state-of-the-art unsupervised GCL counterparts on multiple benchmark graph datasets.

\clearpage
\section*{Acknowledgments}
\thanks{This work is supported by NSF (No. 2229461), ARO (No. W911NF2110088), ONR (No. N00014-21-1-4002), and ARL (No. W911NF2020124).}



\bibliography{neurips}


\clearpage

\appendix
\input{Appendix.tex}
\end{document}

%% file: Appendix.tex
\section{Dataset Details}
\label{app:dataset}

For the experiments on Cora, Citeseer, and Pubmed datasets, we use the same train/validation/test data splits adopted by \cite{kipf2017semi,velickovic_2019_iclr}. For the Amazon-P and Coauthor CS datasets, we follow the same train/validation/test data splits in~\cite{shchur2018pitfalls}. For the experiments with ogbn-arxiv, we follow the OGB benchmarking protocol \cite{hu2020open}. Specifically, for the node classification with few labels task, we sample partial labeled nodes from the original training set and use the same validation and test splits as the standard node classification task. Note that for each of the datasets, we run the experiment 10 times and report the average performance. 

\begin{table}[h]
\small
	\centering
	\begin{tabular}{@{}lcccc@{}}
		\toprule
		\textbf{Dataset} &  \textbf{Nodes} & \textbf{Edges} & \textbf{Features} & \textbf{Classes} \\ \midrule
		\textbf{Cora}              & 2,708    & 5,429        & 1,433            & 7                \\
		\textbf{Citeseer}         & 3,327    & 4,732        & 3,703            & 6                \\
		\textbf{Pubmed}           & 19,717   & 44,338      & 500               & 3                \\
		\textbf{Amazon-P}     & 7,650     & 238,162       & 745              & 8            \\
		\textbf{Coauthor CS}     & 18,333     & 81,894       & 6,805              & 15            \\
		\textbf{ogbn-arxiv}     & 169,343     &  1,166,243       & 128     & 40            \\
 \bottomrule
	\end{tabular}
		\caption{The statistics of the datasets.}
	\label{tab:dataset}
\end{table}

\section{Implementation Details}
\label{app:implementation}

We implement our proposed framework in PyTorch and optimize it with Adam~\cite{kingma_2014_iclr}. All experiments are conducted on a Nvidia Tesla v100 16GB GPU. We set $L$ to $10$ in our multi-scale feature propagation. The output dimension of our encoder network is fixed to $512$. The number of hidden units for the MLP projector is set to 2048. We set the number of negative samples $M$ in $\mathcal{L}_{str}$ to 512. For the training of S$^3$-CL, we first pre-train the encoder network by minimizing the structural contrastive loss $\mathcal{L}_{str}$ and initialize the model parameters with the pre-trained weights. After that, we optimize the model as illustrated in Algorithm \ref{alg:Algorithm-DGCL}. We tune the balancing parameter $\gamma$ within the search space $\{0.2,0.3, 0.4, 0.5, 0.6, 0.7, 0.8\}$. The value of $\xi$ in our Bayesian non-parametric prototype inference algorithm is selected from $\{0.1, 0.15, 0.2, 0.25, 0.3, 0.35, 0.4, 0.45, 0,5\}$. For all the baseline methods, we also use Adam as the optimizer. We do grid search for the learning rate in \{$1 \times 10^{-5}$, $5 \times 10^{-5}$, $1 \times 10^{-4}$, $5 \times 10^{-4}$, $1 \times 10^{-3}$, $5 \times 10^{-3}$, $1 \times 10^{-2}$, $5 \times 10^{-2}$, $1 \times 10^{-1}$, $5 \times 10^{-1}$ \} on different datasets. The temperature parameters $\tau_1$ and $\tau_2$ are set to 1 in our experiments. Our implementation of S$^3$-CL can be found at \url{https://github.com/kaize0409/S-3-CL.}

\vspace{0.1cm}

\subsection{Bayesian Non-parametric Prototype Inference Algorithm}
\label{sec:Algorithm-Proto}

To better illustrate the process of Bayesian non-parametric prototype inference, We summarize the steps in Algorithm~\ref{alg:Algorithm-Proto}.


\subsection{Prototype Refinement via Label Propagation} 
\label{sec:label}
Considering the fact that the pseudo labels inferred by the Bayesian non-parametric algorithm could be inaccurate, we further refine the pseudo labels by label propagation \cite{zhou2004learning} on the graph. This way we can smooth the inaccurate pseudo labels and refine the cluster prototype representations by leveraging graph structure knowledge. Firstly, we convert the prototype assignments $\mathcal{Z}$ to a one-hot pseudo label matrix $\mathbf{Z} \in \mathbb{R}^{N \times K}$, where $\mathbf{Z}_{ik}=1$ if and only if $z_i = k$. Following the idea of Personalized PageRank (PPR)~\cite{klicpera2019predict}, the pseudo labels after $T$ aggregation steps $\mathbf{Z}^{(T)}$ are updated by:
\begin{equation}
\label{eq:propagation}
        \mathbf{Z}^{(t+1)} = (1 - \beta) \Tilde{\mathbf{A}}_{sym} \mathbf{Z}^{(t)} + \beta\mathbf{Z}^{(0)},
\end{equation}
where $\mathbf{Z}^{(0)} = \mathbf{Z}$ and $\beta$ can be considered as the teleport probability in PPR. Next, we convert the propagated results $\mathbf{Z}^{(T)}$ back to hard pseudo labels by setting $z_i = \arg\max_k \mathbf{Z}^{(T)}_{ik}$ for $i\in\{1,...,N\}$.

\begin{algorithm}[h!]
\caption{Algorithm of Bayesian Non-parametric Prototype Inference.}
\label{alg:Algorithm-Proto}
\LinesNumbered
\small
\KwIn{Node representation $\mathbf{h}_1, ..., \mathbf{h}_N$; threshold to generate new prototype $\beta$}
\KwOut{Prototype representation $\mathbf{C}$, pseudo labels $\mathcal{Z}$, and number of clusters $K$}

Initialize $\Tilde{K}=1$, $z_i=1$ and $\mathbf{c}_1 = \frac{1}{N} \sum_{i=1}^{N}\mathbf{h}_i$.

\While{not converge}{

\For{$i = 1$ to $N$}{
    Update the pseudo label $z_i$ according to Eq.(\ref{eq:label_update})

    If $\min_k d_{ik} > \beta$, set $\Tilde{K} = \Tilde{K}+1$
}

\For{$k = 1$ to $\Tilde{K}$}{
    Compute the prototype representation $\mathbf{c}_k = \sum_{z_i=k}\mathbf{h}_i$
}
}

Set $K = \Tilde{K}$

\Return $\mathbf{C}$, $\mathcal{Z}$, and $K$

\end{algorithm}

\section{Additional Experimental Results}
\vspace{0.1cm}

\subsection{Study on Prototype Inference.}
In the proposed semantic contrastive learning, we propose a Bayesian non-parametric algorithm to infer prototypes of node representations. During the training, a new prototype will be instantiated if the distance between all existing prototype representations and the representation of a node is larger than a threshold $\beta$. 

\begin{table}[!htbp]
	\small
	\centering
	\scalebox{1.0}{

\begin{tabular}{cccccc} 
\toprule
Datasets 		& Cora & Citeseer 		 & Pubmed \\ 
\midrule
$\beta$         & 0.20     		 & 0.15       				& 0.35  \\
Estimated $K$   & 8        		 & 6          				& 3     \\
Classes         & 7        		 & 6          				& 3     \\

Baseline        & 84.8$\pm{0.6}$ & 74.6$\pm{0.3}$            & 80.7$\pm{0.2}$  \\
S$^3$-CL        & 84.5$\pm{0.4}$ & 74.6$\pm{0.4}$            & 80.8$\pm{0.3}$  \\
\bottomrule
\end{tabular}

}
	\caption{Ablation study on prototype inference.}
	\label{tab:proto}
\end{table}
\begin{figure*}[htbp]
    
    \graphicspath{{figures/}}
    \centering
    \scalebox{0.9}{
    \hspace{-0cm}
    \subfigure[Cora]
    {
    \includegraphics[width=0.27\textwidth]{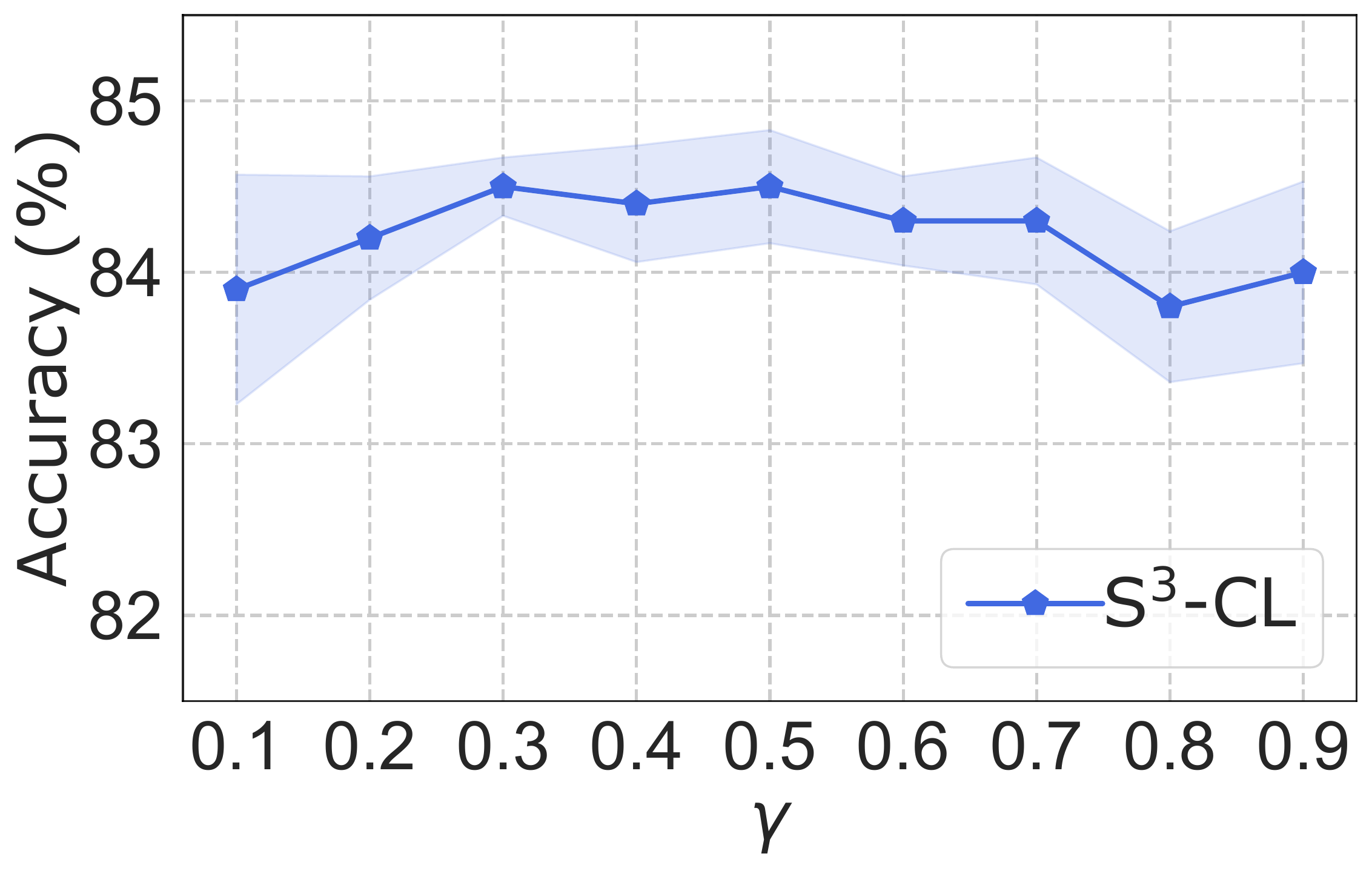}
    }
    \hspace{-0cm}
    \subfigure[Citeseer]
    {
    \includegraphics[width=0.27\textwidth]{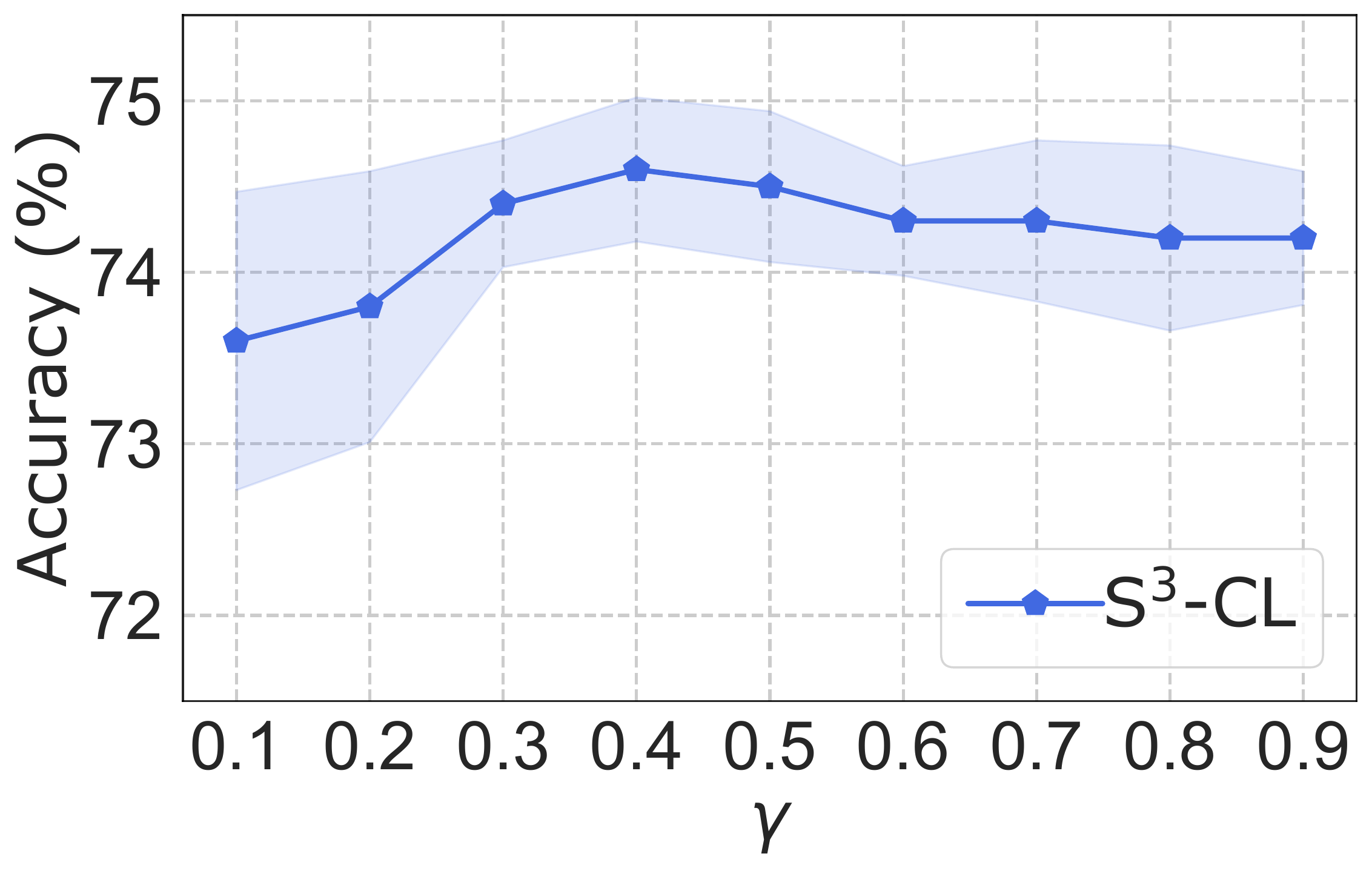}
    }
    \hspace{-0cm}
    \subfigure[Pubmed]
    {
    \includegraphics[width=0.27\textwidth]{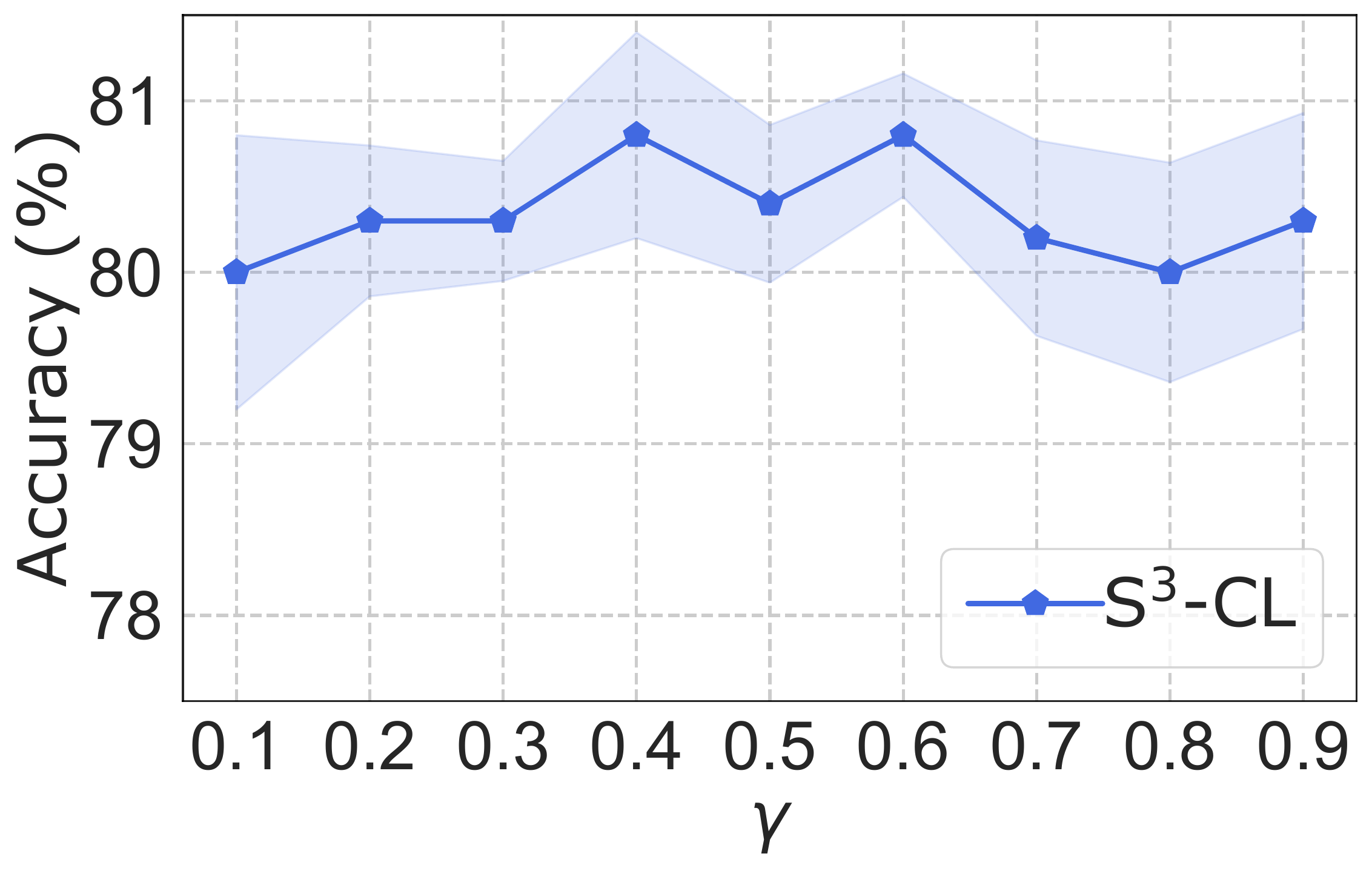}
    }
    }
    \caption{Effect analysis on the value of balancing parameter ($\gamma$).}%
    \label{fig:balance}
\end{figure*}

To verify the effectiveness of the prototype inference algorithm, we design a baseline model that adopts K-Means clustering algorithm to infer prototypes as in \cite{li2020prototypical}. For a fair comparison, label propagation is also applied to refine the prototypes obtained by the baseline model. The results shown in Tabel~\ref{tab:proto} demonstrate that the proposed Bayesian non-parametric prototype inference algorithm is good enough to capture semantic information of the nodes, even without the knowledge on the number of classes. 
\vspace{0.1cm}

\subsection{Effect of Balancing Parameter.}
In this subsection, we study the effect of the balancing parameter $\gamma$ in the final loss function on three datasets, including Cora, Citeseer and Pubmed. Note that we also have similar observations on the other datasets. The results from Figure~\ref{fig:balance} show that S$^3$-CL usually achieve the best results with balancing parameter $\gamma$ set between $0.4$ to $0.6$. 
\vspace{0.1cm}

\subsection{Effect of Negative Example Sampling} 
\label{app:negative}
We study the influence of the size of sampled negative examples  in the structural contrastive loss of S$^3$-CL. The results are shown in Table~\ref{tab:negative}. We can see that as the number of sampled negative examples increase from a small value, the performance of S$^3$-CL can be improved. However, further improving the number of sampled negative examples to $M > 512$ does not lead to better performance. 
\begin{table}[!htbp]
	\small
	\centering
	\scalebox{1.0}{
\begin{tabular}{cccccc} 
\toprule
$M$  & Cora & Citeseer& Pubmed         \\ 
\midrule
$64$ 	& 84.0$\pm$0.2 & 74.1$\pm$ 0.3& 80.3$\pm$0.3 \\
$256$ 	& 84.2$\pm$0.4 & 74.4$\pm$ 0.4& 80.3$\pm$0.2 \\
$512$	& 84.5$\pm$0.4 & 74.6$\pm$ 0.4& 80.8$\pm$0.3 \\ 
$1024$ 	& 84.6$\pm$0.7 & 74.6$\pm$ 0.6& 80.7$\pm$0.6 \\
$2048$ 	& 84.6$\pm$0.5 & 74.5$\pm$ 0.2& 80.8$\pm$0.5 \\
\bottomrule
\end{tabular}
}
	\caption{Effect analysis on the value of balancing parameter ($\gamma$) in the overall loss function.}
	\label{tab:negative}
\end{table}

\begin{table*}[h]
	\centering
	\scalebox{0.75}{
\begin{tabular}{c|ccc|ccc|ccc}
\hline
\multirow{2}{*}{Methods} & \multicolumn{3}{c|}{Cora}                & \multicolumn{3}{c|}{Citeseer}            & \multicolumn{3}{c}{Pubmed}               \\ \cline{2-10} 
                         & Time (s) & Memory (MB) & Params          & Time (s) & Memory (MB) & Params          & Time (s) & Memory (MB) & Params          \\ \hline
DGI                      & 4.2      & 3817        & 7.3$\times10^5$ & 3.5      & 3950        & 1.9$\times10^6$ & 8.4      & 3750        & 2.6$\times10^5$ \\
GMI                      & 6.2      & 4155        & 9.9$\times10^5$ & 4.7      & 4306        & 2.2$\times10^6$ & 10.2     & 4028        & 5.2$\times10^5$ \\
MVGRL                    & 7.6      & 2301        & 9.9$\times10^5$ & 4.4      & 2608        & 2.2$\times10^6$ & 13.6     & 2430        & 5.2$\times10^5$ \\
MERIT                    & 9.2      & 1801        & 7.3$\times10^5$ & 6.5      & 1825        & 1.9$\times10^6$ & 15.2     & 1723        & 2.6$\times10^5$ \\
SUGRL                    & 4.2      & 1610        & 9.7$\times10^5$ & 3.6      & 1754        & 2.6$\times10^6$ & 10.3     & 1740        & 3.9$\times10^5$ \\
S$^3$-CL                 & 5.2      & 1405        & 7.3$\times10^5$ & 3.9      & 1660        & 1.9$\times10^6$ & 12.3     & 1580        & 2.6$\times10^5$ \\ \hline
\end{tabular}
	}
	\caption{Efficiency comparisons between S$^3$-CL and baseline methods w.r.t. training time (seconds/epoch), memory occupation (MB) and number of parameters.}
	\label{tab:time1}
\end{table*}
\begin{table*}[h]
	\centering
	\scalebox{0.75}{
\begin{tabular}{c|ccc|ccc|ccc}
\hline
\multirow{2}{*}{Methods} & \multicolumn{3}{c|}{Cora}                & \multicolumn{3}{c|}{Citeseer}            & \multicolumn{3}{c}{Pubmed}               \\ \cline{2-10} 
                         & Time (s) & Memory (MB) & Params          & Time (s) & Memory (MB) & Params          & Time (s) & Memory (MB) & Params          \\ \hline
DGI                      & 5.9      & 3967        & 3.1$\times10^6$ & 4.5      & 4071        & 3.3$\times10^6$ & 9.9      & 3874        & 2.6$\times10^6$ \\
GMI                      & 7.3      & 4275        & 3.4$\times10^6$ & 5.9      & 4430        & 4.6$\times10^6$ & 12.0     & 4120        & 2.9$\times10^6$ \\
MVGRL                    & 9.2      & 2507        & 3.4$\times10^6$ & 6.5      & 2908        & 4.6$\times10^6$ & 14.8     & 2750        & 2.9$\times10^6$ \\
MERIT                    & 11.0     & 2105        & 3.1$\times10^6$ & 7.8      & 2023        & 3.3$\times10^6$ & 16.3     & 2010        & 2.6$\times10^6$ \\
SUGRL                    & 5.5      & 1820        & 1.2$\times10^6$ & 4.7      & 1904        & 2.8$\times10^6$ & 11.9     & 1867        & 5.4$\times10^5$ \\
S$^3$-CL                 & 5.2      & 1405        & 7.3$\times10^5$ & 3.9      & 1660        & 1.9$\times10^6$ & 12.3     & 1580        & 2.6$\times10^5$ \\ \hline
\end{tabular}
	}
	\caption{Efficiency comparisons between S$^3$-CL and baseline methods (10-layer encoder network) w.r.t. training time (seconds/epoch), memory occupation (MB) and number of parameters.}
	\label{tab:time_10}
\end{table*}
\vspace{0.1cm}

\subsection{Efficiency Analysis.}
\label{sec:efficiency}
In this subsection, we evaluate the model efficiency of different methods in terms of training time, memory consumption, and parameter size. We first compare S$^3$-CL with existing unsupervised GCL methods and show the results in Table \ref{tab:time1}. Though the baseline methods only use shallow GNNs that cannot capture global structural knowledge, S$^3$-CL still has less memory consumption and smaller parameter size, and also is competitive with the most efficient baselines DGI and SUGRL in terms of training time per epoch. The main reason is that our approach uses multi-scale feature propagation to capture the global structural knowledge, which saves the augmentation time during the training and allows the network encoder to be built with one-layer neural network.

Moreover, for each of the existing unsupervised GCL methods, we increase the encoder network depth to $10$ and show their efficiency evaluation results in Table \ref{tab:time_10}. It is noteworthy that by increasing the depth of the GNN encoder, existing methods have to introduce more model parameters to capture long-range node interactions, leading to slower training speed and larger memory consumption. In this case, our proposed framework S$^3$-CL has the best efficiency for capturing the same scale of global structural knowledge. 
\vspace{0.1cm}

\subsection{Effect Analysis on temperature parameters.}
In this subsection, we examine the effect of the temperature parameters $\tau_1$ and $\tau_2$. Specifically, we vary the value between \{0.2, 0.4, 0.6, 0.8, 1\} for each temperature parameter, and report the results in Table~\ref{tab:tau1} and Table~\ref{tab:tau2}. It is observed that S$^3$-CL generally performs better when the temperature parameters are close to $1$.



\begin{figure*}[htbp!]
\begin{floatrow}
\capbtabbox{%

\scalebox{1}{

\begin{tabular}{cccc}
\hline
$\tau_1$ & Cora         & Citeseer      & Pubmed       \\ \hline
$1$      & 84.5$\pm$0.4 & 74.6$\pm$ 0.4 & 80.8$\pm$0.3 \\
$0.8$    & 84.4$\pm$0.4 & 74.6$\pm$ 0.4 & 80.7$\pm$0.2 \\
$0.6$    & 84.5$\pm$0.4 & 74.3$\pm$ 0.6 & 80.6$\pm$0.4 \\
$0.4$    & 84.4$\pm$0.6 & 74.6$\pm$ 0.3 & 80.5$\pm$0.6 \\
$0.2$    & 84.1$\pm$0.5 & 74.4$\pm$ 0.3 & 80.4$\pm$0.5 \\ \hline
\end{tabular}
    
}
}{%
  \caption{Effect analysis on the value of $\tau_1$.}%
  \label{tab:tau1}
}
\capbtabbox{%

\scalebox{1}{

\begin{tabular}{cccc}
\hline
$\tau_1$ & Cora         & Citeseer      & Pubmed       \\ \hline
$1$      & 84.5$\pm$0.4 & 74.6$\pm$ 0.4 & 80.8$\pm$0.3 \\
$0.8$    & 84.3$\pm$0.5 & 74.6$\pm$ 0.4 & 80.7$\pm$0.3 \\
$0.6$    & 84.4$\pm$0.3 & 74.3$\pm$ 0.5 & 80.8$\pm$0.4 \\
$0.4$    & 84.2$\pm$0.6 & 74.6$\pm$ 0.4 & 80.3$\pm$0.7 \\
$0.2$    & 84.2$\pm$0.5 & 74.3$\pm$ 0.6 & 80.2$\pm$0.6 \\ \hline
\end{tabular}
    
}
}{%
  \caption{Effect analysis on the value of $\tau_2$.}%
  \label{tab:tau2}
}
\end{floatrow}
\end{figure*}

\vspace{0.3cm}
\section{Additional Related Work}
\label{sec:add-related}

\textbf{Deep Clustering.}  Clustering-based unsupervised representation learning is of particular interest recently. Various methods are proposed to jointly optimize feature learning and image clustering. Deep Embedding Clustering
(DEC) \cite{huang2014deep} learns a mapping from the data space
to a lower-dimensional feature space, in which, it iteratively optimizes a clustering objective. DeepCluster \cite{Caron_2018_ECCV} uses k-means to assign pseudo-labels to learn visual representations. This method scales to large uncurated dataset and can be used for pre-training of supervised networks. Following that, DeeperCluster~\cite{caron2019unsupervised} is proposed to combine the objective of self-supervised learning and clustering. SwAV~\cite{caron2020unsupervised} further improves this idea by simultaneously clustering the data while enforcing consistency between cluster assignments produced for different views. SCL~\cite{huang2021deep} also seeks to impose the semantic information into the unlabelled training data through a clustering objective. Recently, a similar work that learns global semantic clustering information is also proposed for graph classification \cite{xu2021self}. However, previous clustering-based self-supervised learning methods adopt a preset number of clusters. In contrast, we propose a Bayesian non-parametric prototype inference algorithm in our semantic contrastive learning to learn the number of clusters and the representation of cluster prototypes at the same time. 